\documentclass[preprint,3p,times,review,12pt]{elsarticle}

\usepackage{amssymb}
\usepackage{amsthm}
\usepackage{amsmath}
\usepackage{gensymb}
\usepackage{booktabs}
\usepackage{multirow}
\usepackage{array}
\usepackage{lineno}
\usepackage{todonotes}
\usepackage{hyperref}
\usepackage[acronym,nomain, automake]{glossaries}
\makeglossaries
\newacronym{CV}{CV}{Computer Vision}
\newacronym{XR}{XR}{Extended Realities}
\newacronym{UI}{UI}{User Interface}
\newacronym{AC}{AC}{Augmented Carpentry}
\newacronym{ACEngine}{ACE}{Augmented Carpentry Engine}
\newacronym{ACE}{ACE}{Augmented Carpentry Engine}
\newacronym{NUC}{NUC}{Intel\textsuperscript{\textregistered} Next Unit of Computing}
\newacronym{SLAM}{SLAM}{Simultaneous Localization and Mapping}
\newacronym{6DoF}{6DoF}{Six Degrees of Freedom}
\newacronym{CDM}{CDM}{Camrea-Display Module}
\newacronym{FoV}{FoV}{Field of View}
\newacronym{ACIM}{ACIM}{Augmented Carpentry Informed Model}
\newacronym{ACIT}{ACIT}{Augmented Carpentry Informed Toolhead}
\newacronym{CNC}{CNC}{Computer Numerical Control}
\newacronym{GO}{GO}{Geometric Object}
\newacronym{FPS}{FPS}{Frame per Second}
\newacronym{PoI}{PoI}{Point of Interest}
\newacronym{SMEs}{SMEs}{Small and Medium-sized Enterprises}
\newacronym{FSO}{FSO}{Federal Statistical Office}
\newacronym{HMD}{HMD}{Head Mounted Display}
\newacronym{TCP}{TCP}{Tool Control Point}
\newacronym{HRC}{HRC}{Human Robot Collaboration}
\newacronym{HCI}{HCI}{Human Computer Interaction}
\newacronym{IMU}{IMU}{Inertial Measurement Unit}
\newacronym{IR}{IR}{Infrared}
\newacronym{BIM}{BIM}{Building Information Modeling}
\newacronym{IO}{IO}{Input/Output}
\newacronym{CAD}{CAD}{Computer-aided Design}

\journal{Automation in Construction}

\begin{document}

\begin{frontmatter}

\title{Augmented Carpentry: Computer Vision-assisted Framework for Manual Fabrication}

\author[inst1]{Andrea Settimi\corref{cor1}}

\affiliation[inst1]{organization={Ecole Polytechnique Fédérale de Lausanne (EPFL), School of Architecture, Civil and Environmental Engineering (ENAC), Institute of Civil Engineering (IIC), Laboratory for Timber Constructions (IBOIS)},
            addressline={GC H2 711, Station 18}, 
            city={Lausanne},
            postcode={1015}, 
            state={Vaud},
            country={Switzerland}}
            \ead{andrea.settimi@epfl.ch}
\affiliation[inst2]{Independent Researcher}

\author[inst2]{Julien Gamerro}
\author[inst1]{Yves Weinand}

\cortext[cor1]{Corresponding author}

\begin{abstract}
Ordinary electric woodworking tools are integrated into a multiple-object-aware augmented framework to assist operators in fabrication tasks. This study presents an advanced evaluation of the developed open-source fabrication software Augmented Carpentry (AC), focusing on the technical challenges, potential bottlenecks, and precision of the proposed system, which is designed to recognize both objects and tools. In the workflow, computer vision tools and sensors implement inside-out tracking techniques for the retrofitting tools. This method enables operators to perform precise saw-cutting and drilling tasks using computer-generated feedback. In the design and manufacturing process pipeline, manual fabrication tasks are performed directly from the computer-aided design environment, as computer numerical control machines are widely used in the timber construction industry. Traditional non-digital methods employing execution drawings, markings, and jigs can now be replaced, and manual labor can be directly integrated into the digital value chain. First, this paper introduces the developed methodology and explains its devices and functional phases in detail. Second, the fabrication methodology is evaluated by experimentally scanning the produced one-to-one scale mock-up elements and comparing the discrepancies with their respective three-dimensional execution models. Finally, improvements and limitations in the tool-aware fabrication process, as well as the potential impact of  AC in the digital timber fabrication landscape, are discussed.
  \end{abstract}




\end{frontmatter}


\section{Introduction}
\label{sec::intro}

\subsection{Context}
\label{sec::intro::context}

The sustainability of timber construction extends beyond the choice of materials. It also depends on the efficiency and adaptability of the production methods. When used effectively, digital fabrication can bridge the gap between the ecological advantages of timber and its scalable implementation. However, the current digital timber fabrication model often relies on centralized, high-tech manufacturing systems.

Although these systems achieve impressive precision and speed, they prioritize repeatability over flexibility and local customization. Their high entry costs ~\cite{He2021}, disruptive nature~\cite{He2021}, maintenance~\cite{Kuzmenko2021}, and technical complexities~\cite{Ng2022} further restrict their use to well-funded firms and research institutions. Consequently, more accessible and flexible digital production approaches remain underexplored, limiting broader community engagement and local digital innovation in timber construction, especially in small contexts where reliance on human labor prevails. In Switzerland, \gls{SMEs} represent the backbone of construction, accounting for over 99\% of all construction businesses and providing approximately two-thirds of the nation's jobs in this sector~\cite{SECO2024}. According to the \gls{FSO}, in 2022, \gls{SMEs} in the construction sector represented 299'000 jobs compared with 34'000 generated by large enterprises with more than 250 employees~\cite{Bfs2024}.

To increase the adoption of digital fabrication technologies in this scenario, a great deal of the current research has focused on making existing high-end robotic systems more accessible and adaptable, primarily through improved interfaces and \gls{HRC} strategies that incorporate human input into robotic sequencing~\cite{Ron2024}. Although this movement towards HRC is often described as a step forward in automation, it also highlights the persistent challenges of achieving such a level of dependence~\cite{Yang2024}. These interface designs and ergonomic efforts have yielded valuable proofs of concept~\cite{Pedersen2021, Amtsberg2021}; however, they are fundamentally based on the reliance on robotic infrastructure. This situation invites discussion on whether digital fabrication should remain primarily oriented towards a select group of firms focusing on high-end hardware integration or broaden its scope to ensure more inclusive accessibility and wider technology dissemination for digital timber construction.

Moreover, the scarcity of semiconductors and rare materials indicates that relying solely on new technology-intensive solutions can become costly and logistically challenging at this scale~\cite{EC2023}. Consequently, strategies seeking to retrofit or adapt existing assets have gained traction as more resilient and resource-efficient alternatives~\cite{Pietrangeli2023}. Retrofitting leverages the established infrastructures and manual workflows, reducing the need for large-scale, high-end electronics, while integrating digital enhancements where feasible. By selectively incorporating upgraded components or redeploying non-specialized hardware, companies can mitigate risks while extending their digital capabilities because of low investment.
Overall, retrofitting is a forward-looking strategy that combines sustainability, cost-effectiveness, and technological evolution. The construction industry can access new digital fabrication frameworks without becoming disproportionately hardware dependent by prioritizing the life extension of existing assets.

Based on these considerations, the principles of retrofitting are particularly relevant when applied to the portable machines that are fundamental to timber construction. These manual tools remain indispensable in various scenarios, from full-scale fabrication and custom piece production to on-site adjustments in \gls{SMEs} and larger enterprises. As the industry navigates material shortages and seeks to digitize existing processes, augmenting existing woodworking power tools has emerged as a natural extension of retrofitting strategies. This approach not only preserves the machinery hardware and operator efforts, but also opens pathways for incorporating advanced technologies in a manner that is cost effectively, maintainable, scalable, and non-disruptive to the pre-existing infrastructure while granting the company access to custom digital fabrication.

Recent advances in computer sensing and tracking technologies have illustrated the feasibility of this approach in several case studies, where both machine functionalities and human performance were effectively enhanced via tool augmentation. However, numerous technical and conceptual challenges have impeded the efficient large-scale deployment of these innovations. The following review synthesizes the key scientific literature on tool retrofitting and augmentation, focusing on timber as the target material, and critically examines the limitations that this work seeks to address.

\subsection{Related works}
\label{sec::intro::relatedworks}

In the broader \gls{HCI} field, intelligent building tools occupy the middle ground between purely manual implements (e.g., hammers) and fully autonomous industrial robots. Equipped with sensing technologies, microcontrollers, and onboard computational power, they  deliver real-time feedback and cognitive and visual support to the user, while preserving the portability, affordability, and intuitive handling of traditional power or manual tools. This section surveys the key research on retrofitting or designing new tools that digitally augment human capabilities with computational feedback, with a particular focus on drilling, fastening, and sawing tasks in timber construction.

The work of Rivers et al. serves as the foundation for the current study, bridging digital precision with manual craftsmanship through tool augmentation and sensor-localization strategies. In their early projector-camera system, Rivers et al. \cite{Rivers2012} demonstrated how unskilled users iteratively shaped precise replicas of digital models by adding or removing material under real-time projected guidance. Building on this seamless integration of computational feedback and human control, Rivers et al. \cite{RiversProtoShaper2012} introduced a router prototype featuring a self-correcting spindle- and  tag-based tracking system—a precursor to what would eventually become the Shaper Origin\textsuperscript{\textregistered}~\cite{ShaperTool2025}, the first commercial computer-guided woodworking device for two-dimensional (2D) fabrication. Although the Shaper Origin\textsuperscript{\textregistered} shows how sensors, visual feedback, and robotic corrections enhance manual processes, it remains limited to navigating 2D paths and lacks true awareness of the borders or shape of the workpiece. Its robust tag-based tracking system enables the cutting of portions of the timber without losing signals, maintaining the execution locked in; however, its dimensional constraints make it unsuitable for large-scale components.

With similar goals, Tian et al. \cite{Tian2018} repurposed an ordinary drill with readily available open hardware to create a portable milling station, although the system still followed the same kinematic principles as traditional multi-axis \gls{CNC} machines and lacked advanced sensing. A subsequent study by Tian et al. \cite{Tian2021} employed a robotic setup, together with the method of Rivers et al.~\cite{RiversProtoShaper2012}, to demonstrate how physical augmentation can physically refine the positioning of a user to near-robotic precision. Despite the high cost, these studies indicate an intriguing future for wearable robotics and physical augmentation in manual tasks with tools.

Haptic guidance coupled with inertial tracking has also been explored as a possible layout for smart retrofitting tools. Zoran et al. \cite{Zoran2014} presented an early demonstration of motion-sensing-based hand-tool tracking, halting the milling motor whenever the monitored bit intersects the target digital model. Subsequently, Tashiro et al. \cite{Tashiro2024} introduced two novel haptic devices that apply non-vibratory tactile feedback to reflect edge interactions in real time. One device, a drill augmented with a compact \gls{IMU} and linear actuators, provides immediate rotational guidance during construction tasks. Despite its ergonomic advantages and simplicity, this study highlights the inherent limitations of purely non-visual systems for an end-to-end design-to-production pipeline.

Millimetric tool-tracking accuracy has often been explored in other controlled scenarios using outside-in monitoring systems that rely on \gls{IR} markers with multiple motion cameras \cite{Kuzhagaliyev2018,Brugnaro2018,Iqbal2024}. Although effective, these setups require installation of external tracking devices, which is impractical in unpredictable construction environments. An upgrade in the use of \gls{IR} was demonstrated by Kunz et al.~\cite{Kunz2020} by substituting an exterior stationary tracking system using the raw feed of the \gls{HMD} Hololens1\textsuperscript{\textregistered}. This approach is not ideal for construction applications and sites because it can obstruct the view of the operator, is subject to lighting variances, requires delicate calibration of the position in relation to the tool, and requires maintenance of the marker within the field of vision of the head \cite{Kriechling2020,Wu2017,Liu2018}.

In our previous work~\cite{Settimi2022}, we explored visual sensing and guidance for retrofitted woodworking tools by registering a stereo camera on a drill to enable visual-inertial odometry. The results indicated that the drill and sensor positions require careful calibration and are susceptible to tool impact and drifting events during generic camera self-localization. Moreover, the execution model locking relies on a QR code that is detected through a headset, along with the inherent imprecision and drift of the employed Hololens2\textsuperscript{\textregistered}. Schoop et al. explored the advantages of basic laser-based sensing microcontroller systems for retrofitting manual tools in multiple applications, notably, by integrating range finders and laser-projected overlays to enhance partial saw cutting with a miter saw and portable drilling~\cite{Schoop2016}. Fazel et al. \cite{Fazel2024} merged these two approaches and embedded a laser and an onboard gyroscope clipped to a hand drill and nail gun by employing a Hololens2\textsuperscript{\textregistered} \gls{HMD}. The system detects the tool center point (\gls{TCP}) of the drill bit by analyzing the red laser pattern in the RGB camera feed of the \gls{HMD}, thereby delivering a robust method for tracking both the \gls{TCP} position and tool rotation. It also proposes a design-to-production pipeline with Fologram\textsuperscript{\textregistered}\cite{Fologram2025}. Despite these innovations, this approach still lacks a robust and consistent method for locking the execution model onto the timber piece, even when the model is small, relying instead on manual hologram alignment without computational feedback or refinement.

Although the literature on applied scenarios shows that a single tool (often a drill) can be tracked with high accuracy, the main technical challenge is to incorporate these tracking data into a robust system that supports multiple tools, detects the timber execution model, and securely locks an execution model onto it.
Accurate tracking and model locking to the physical piece remain fundamentally unresolved challenges in the domain of \gls{CV}-enhanced digital construction, including our scenario. Prior work typically considered the localization and alignment of the timber piece as a separate detection from camera or headset self-localization \cite{Parry2021,Gwyllim2024}, introducing additional tolerances that reduced the overall accuracy. Moreover, this approach severely restricts the feasible length of the timber elements, thereby limiting the applicability of the technology to common beam sizes. Although these tolerances can be mitigated by adding multiple fiducial markers along the timber piece, each marker must be calibrated precisely to the three-dimensional (3D) model \cite{Kyaw2023}.
One of the most advanced and promising approaches for object and camera localization in construction tasks was introduced by Sandy et al. \cite{Sandy2016}, emphasizing the object rather than its surroundings. This method has evolved into a markerless self-localization system for monocular cameras \cite{Sandy2018} that relies on object contours rather than background landmarks. Therefore, it provides robust sensor self-localization, even when the target object is moved or repositioned. However, this approach encounters multiple challenges in woodworking, particularly for beams. As the detection depends on a fixed 3D model, the shape of the beam changes as joineries are added, making it difficult to track modifications and precisely determine the new position of the sensor. Moreover, the sensor operates at a close range to monitor both the camera and tool, often causing distortion. Dust, wood chips, and vibrations further compound this problem by introducing additional clutter and noise, which complicates the tracking process.

\subsection{Objectives}
\label{sec::intro::obj}

This study introduces an augmented framework named \textit{\gls{AC}}, which intends to replace traditional 2D execution drawings, manual markings, and physical jigs in timber fabrication with a digital visual guidance system that is compatible with cutting and drilling woodworking electric tools. It leverages affordable sensors, a dedicated \gls{XR} engine, custom \gls{CV} components, a familiar interface, and conventional manual equipment to achieve tool guidance and execution model visualization for timber elements of common construction lengths. We illustrate how tailored \gls{CV}components provide real-time data, continuous metrics, and digital precision drilling and cutting guidance, thereby reducing reliance on exhaustive drawings and manual marking.

This paper details the methodology and operational phases that integrate hardware and software components to guide complex woodworking tasks using computer-processed feedback. A working prototype is presented and validated via an experimental campaign using 3D scanning to assess the accuracy of the fabricated structures against their corresponding 3D execution models. Finally, the potential improvements and new perspectives for \gls{AC} are discussed, underscoring its potential to democratize digital timber fabrication on a distributed, affordable, and regional scale.

\section{Developed methodology}
\label{sec::devmethod}

\textit{AC} uses a purpose-built \gls{XR} engine that is tailored to address the specific fabrication requirements of the presented methodology: \gls{ACEngine}, a lightweight, open-source C++ engine enabling rapid prototyping on UNIX 64-bit machines with a particular focus on \gls{CV}-guided fabrication. In fact, current commercial \gls{XR} engines are widely adopted for such scenarios, but they also exhibit several limitations, including excessive feature sets and proprietary and hardware constraints, but mostly Windows-centric environments, creating barriers for the seamless integration of advanced robotic vision algorithms, which are typically developed in low-level languages for UNIX platforms and are necessary to craft accurate detection solutions in construction. To overcome these limitations, \gls{ACEngine} is structured as a modular and scalable layer-stack flow that is customizable via a single top-level application file and supports the incorporation of touch displays, monocular sensors, and robotic vision components for real-time camera processing, pose estimation, and object tracking. A robust geometric framework unifies the management of 3D objects encompassing \gls{CAD} execution models and tool-specific feedback cues, whereas a computed feedback system delivers specialized visual guidance for tasks such as drilling and cutting. By leveraging a simple monocular camera feed, \gls{ACEngine} overlays instructions by rendering virtual object projections in real time using the OpenGL framework ~\cite{OpenGL1999}. With minimal dependencies, a lean, bloat-free \gls{UI} system ~\cite{Cornut2023}, and complete Linux compatibility, \gls{ACEngine} addresses this critical gap by providing a streamlined pathway for integrating advanced robotic \gls{CV}components into open-source \gls{CV}-assisted construction solutions.

\begin{figure}[!ht]
  \centering
  \includegraphics[width=160mm]{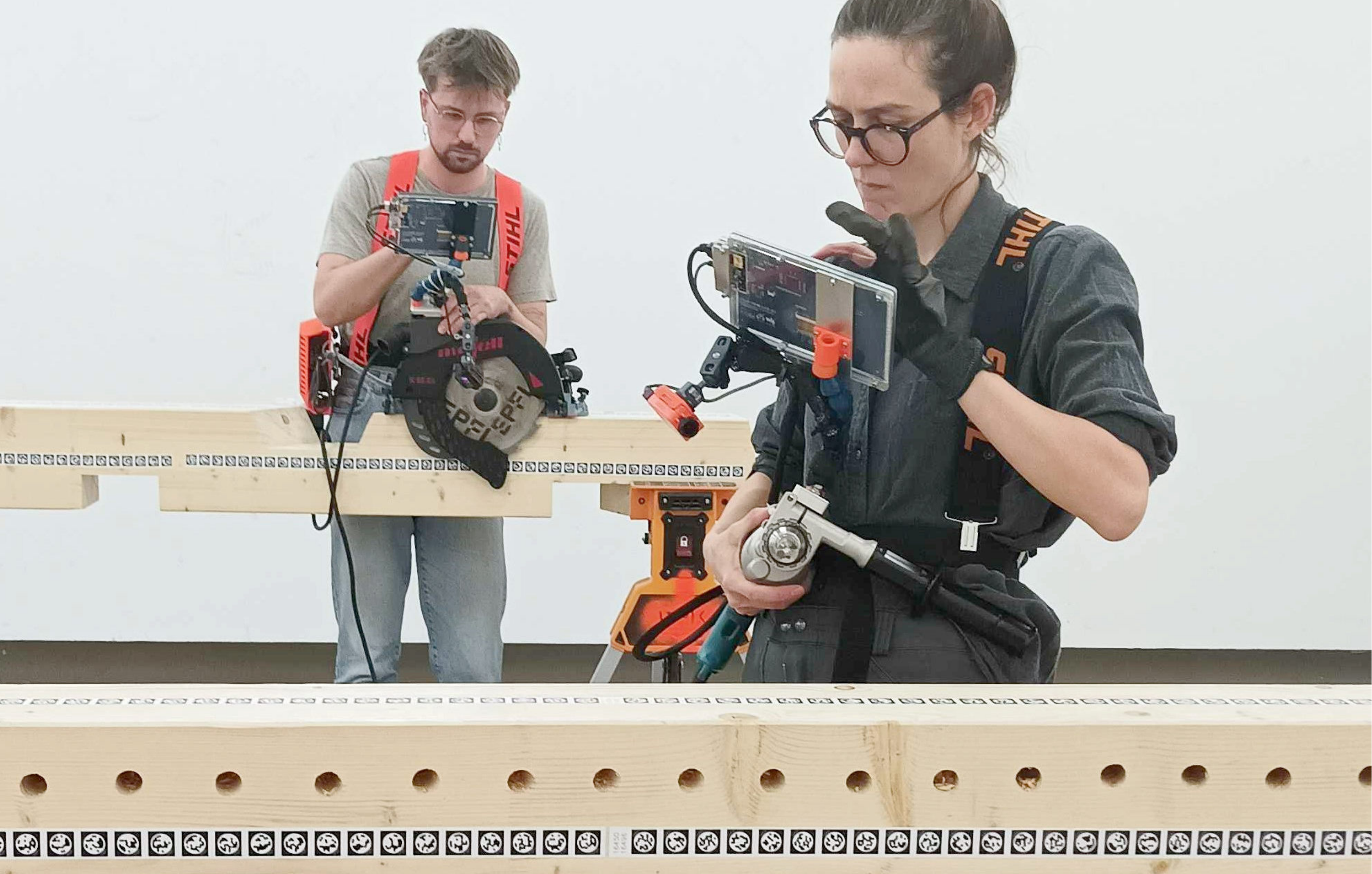}
  \caption{Users experienced the proposed AR-based guiding system for enhanced woodworking during an organized public workshop. Shown are a user utilizing an \textit{AC} setup for saw cutting (back view) and another interacting with a touch display for augmented-guided drilling (front view).}\label{fig:dedvmethod:introac}
\end{figure}

Powered by such an engine, \gls{AC} is designed to unfold in three distinct phases. First, the power tools are equipped with onboard sensors and interfaces, and their tool heads are integrated into an augmented framework. Second, the timber stock is mapped and the corresponding 3D shop drawings are generated. Finally, the AR -guided operations can be performed (Fig.~\ref{fig:dedvmethod:introac}).

\subsection{Tool-augmented retrofitting}
\label{sec::devmethod::retrofit}

\begin{figure}[!ht]
  \centering
  \includegraphics[width=90mm]{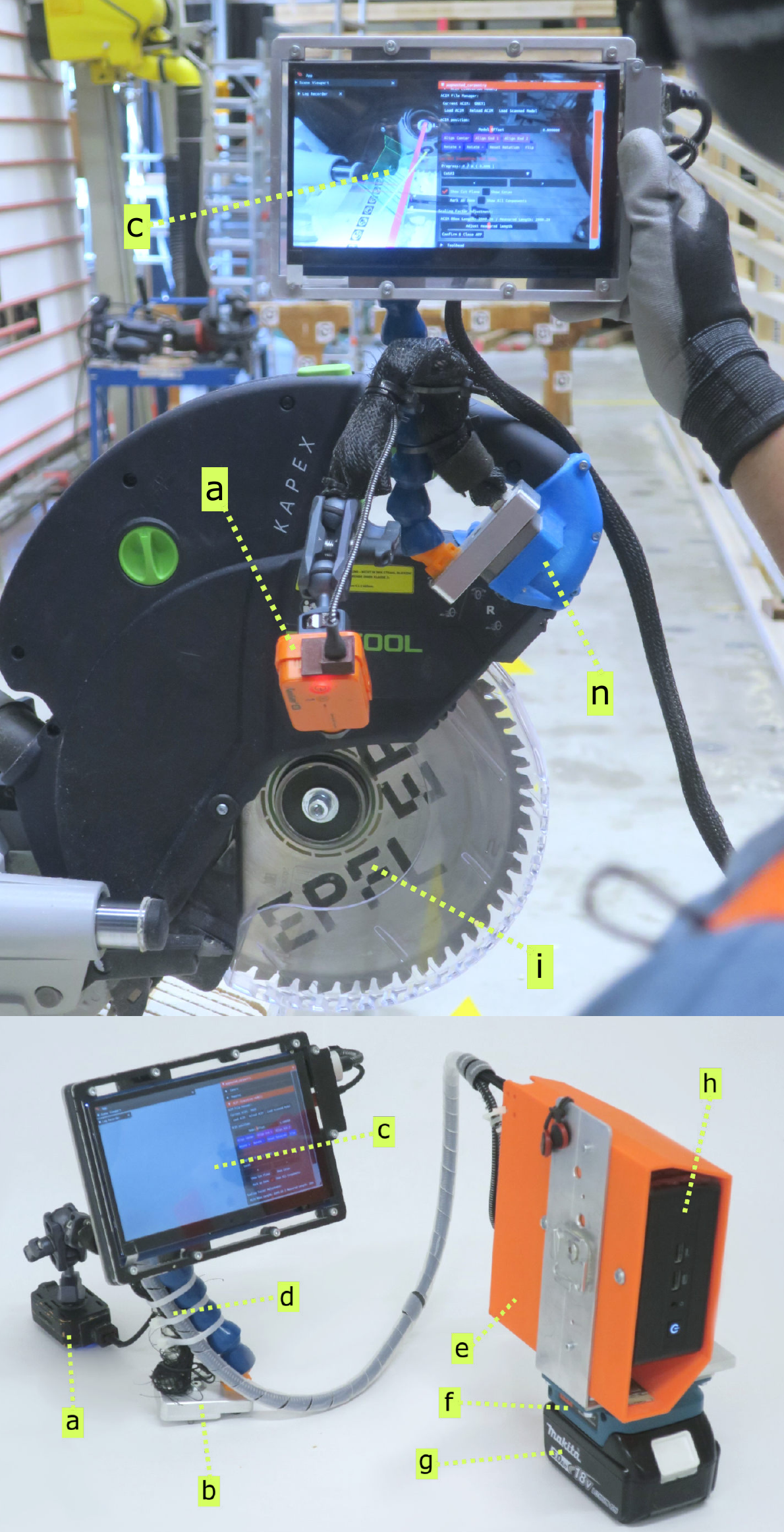}
  \caption{Overview of the designed hardware set-up mounted on a miter saw and once detached: (a) the monocular RGB camera, (b) magnetic clip for the onboard sensor and interface, (c) touch display, (d) two articulated arms for the camera and display, (e) protective case with belt clip, (f) battery adapter, (g) power tool 18V battery, (h) \gls{NUC}, (i) mounted tool head, (n) 3D-printed power tool's mount adapted to each power tool model.}\label{fig:dedvmethod:harddetailon}
\end{figure}

This section details the retrofitting procedures used to integrate woodworking power tools into the \gls{AC} framework. The hardware setup of \gls{AC} is composed of three primary devices: (1) a 7~inch mounted touch-display as the main \gls{IO} device (Fig.~\ref{fig:dedvmethod:harddetailon}(c)), (2) an onboard, fish-eye RGB monocular camera, RunCam2 (1080p) (Fig.~\ref{fig:dedvmethod:harddetailon}(a)) pointing towards the tool head, and (3) a portable 64-bit Intel\textsuperscript{\tiny\textregistered} \gls{NUC} running \gls{AC} (Fig.~\ref{fig:dedvmethod:harddetailon}(h)) that collects the camera feed, processes the data, and generates visual cues to guide the operator in correctly performing the cutting or drilling tasks.

While the user wears the computing unit, the \gls{CDM} can be easily swapped among different tools owing to the magnetic mounting system between the sensor and tools, ensuring interchangeability of the \gls{CDM} toolset (Fig. ~\ref{fig:dedvmethod:harddetailon}. b,n). Therefore, each tool to be retrofitted requires a dedicated 3D-printed mount. All mounts that are produced, including design files and documentation, are available in an open-access hardware repository~\cite{ACMountsRepo2024}.

We selected a simple 7~inch touch display for the interface. This choice simplifies the development process, unlike current \gls{HMD}s, which require self-localization within the environment, thereby introducing additional tolerance constraints. This also results in a streamlined, familiar interface with a minimal learning curve that can be easily mounted on each tool. Concerning the \gls{UI} design, only essential elements are presented to ensure ease of use. The panel on the right contains collapsible subpanels organized by the area of interest. Each subpanel provides widgets that can be manipulated directly using a touch monitor. The remainder of the screen is occupied by an undistorted, 30 FPS grayscale camera feed overlaid solely with color-coded augmented visual cues, as illustrated in Fig. ~\ref{fig:dedvmethod:harddetailon}.

An ordinary monocular camera was selected as a unique sensor for the system. First, it is affordable, widely available, and lightweight (49 g). Second, its wide \gls{FoV} (170°) conveniently captures both the tool head and nearby timber sections. Third, cameras designed for drone deployment endure constant vibrations and shocks produced by the tool, making them well-suited for onboard mounting. Finally, it readily captures both near and distant objects simultaneously, which is a key advantage over range-finding or laser-based systems that often struggle to capture close subjects.

\begin{figure}[!ht]
  \centering
  \includegraphics[width=120mm]{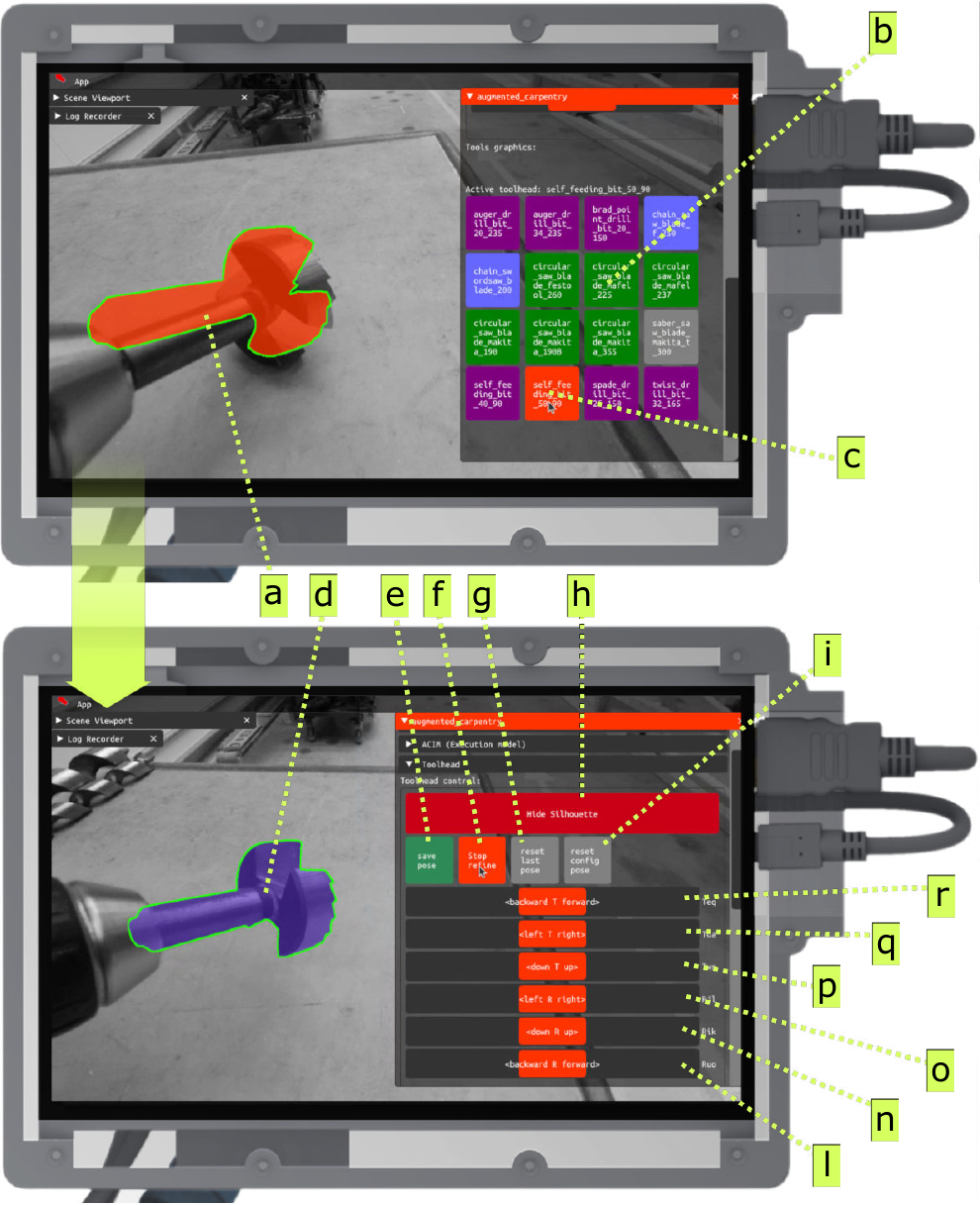}
  \caption{\gls{UI} resuming the controls to input and refine the toolhead \gls{6DoF} pose: (a) rough projection of the selected toolhead, (b) library of selectable toolhead models, (c) currently selected toolhead, (d) machine-refined toolhead position and orientation, (e) confirming the pose, (f) stopping the refiner, (g) reloading the latest saved pose, (h) hiding the 3D model silhouette widget, and (i) resetting the pose to the default value. The user can input the 3D model transformation matrix single values by interacting with six sliders, each representing the model translation axis x (l), y (r), z (p) or the object rotation through axis x (l), y (o), or z (n).}\label{fig:dedvmethod:calibttool}
\end{figure}

\begin{figure}[!ht]
  \centering
  \includegraphics[width=140mm]{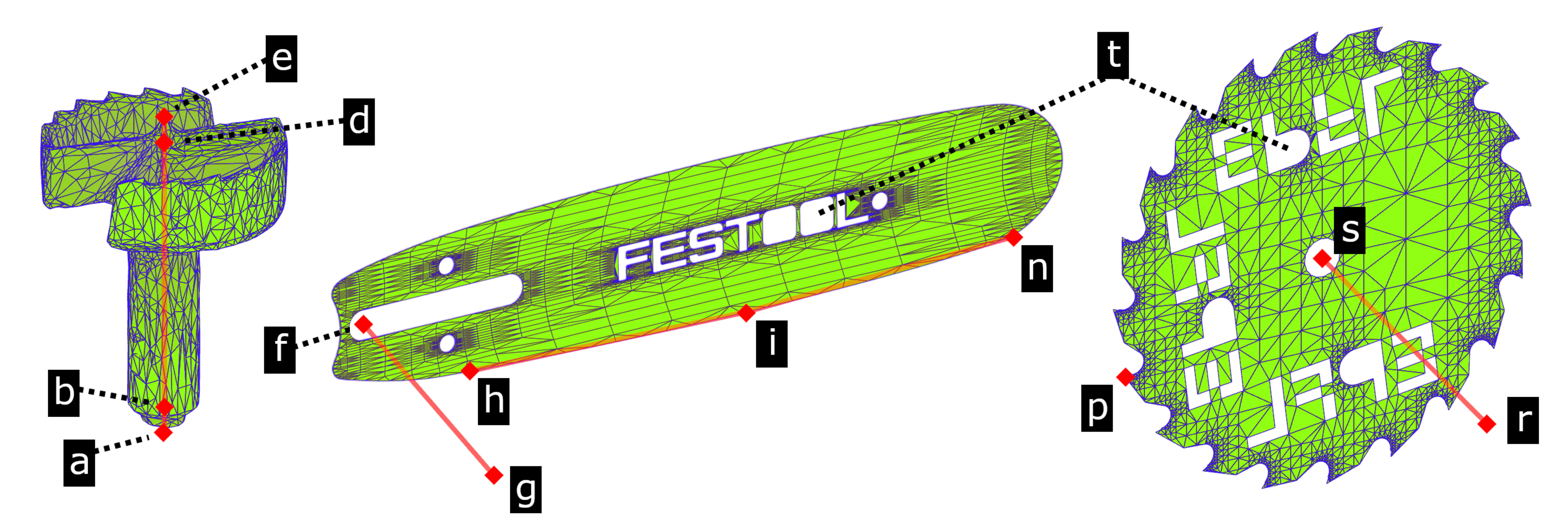}
  \caption{\gls{PoI}s embedded in \gls{ACIT} files loaded together with each entry of drill bits, chainsaw bars, and circular saw blades in the toolhead library of \gls{AC}: drill bit (a) base, (b) chuck, (d) eating, and (e) tool tips; chainsaw bar (f) base, (g) normal, chainsaw (h) start, mid and end points; circular saw blade (i) center, (r) normal and (p) radial point. It is worth noting that (t) labels are incorporated as edges to supply additional edge data for the \gls{6DoF} pose detector when possible.}\label{fig:dedvmethod:acit}
\end{figure}

The aim of embedding the tools with the presented setup is to sense the 3D location and orientation of all relevant tools involved in the augmentation process. Notably, \gls{AC} focuses only on tracking the tool head (e.g., the blade or drill bit), rather than the entire tool. This avoids potential issues arising from different or defective machine models, thereby supporting the seamless integration of legacy and new power tools. To this end, \gls{AC} incorporates a real-time \gls{6DoF} toolhead detector, TTool~\cite{TTool2024}, which computes the pose of a toolhead in space from the onboard monocular camera feed, provided that an accurate 3D mesh of the toolhead is available. All existing toolheads are uploaded to an open-access dataset~\cite{TToolDataset2024} and loaded at runtime (Fig. ~\ref{fig:dedvmethod:acit}). Similar to most \gls{6DoF} detection systems, TTool follows a two-phase approach: (i) acquisition of the global pose and (ii) subsequent refinement. In the first phase, \gls{AC} leverages user expertise by allowing the operator to select a toolhead 3D model from a library (Fig.~\ref{fig:dedvmethod:calibttool}(b)) and approximately align its projection to the physical silhouette via the touch display (Fig.~\ref{fig:dedvmethod:calibttool}(l-r)). The integrated detector then performs refinement based mostly on a contour comparison of this initial estimation to achieve high precision (Fig.~\ref{fig:dedvmethod:calibttool}(d)), transforming the 3D replica in the digital environment of \gls{AC} such that it closely mirrors its physical counterpart. The calibration process is repeated only when the tool or toolhead is changed or the camera position is altered. Notably, continuous tracking is not required during the augmented woodworking process, even if the toolhead is partially occluded (e.g., when cutting material), because of the fixed position of the camera relative to the toolhead.

\subsection{Timber stock acquisition}
\label{sec::devmethod::timberstck}

In any augmentation process, the precise localization of timber is closely tied to the selected strategy for camera self-localization. This principle underpins all \gls{XR} applications that rely on accurately mirroring the camera position in real and virtual spaces to ensure that the execution model aligns with its physical counterpart before being displayed to the user. Achieving such precision becomes very challenging in a woodworking context for several reasons: (i) the shape of the timber element can change throughout the process because of sawing and drilling, (ii) the scene is often cluttered with chips and dust, (iii) high accuracy is required to guide these operations via augmentation (in the order of 1~mm), and (iv) the entire timber shape may not always be fully present in the camera \gls{FoV}.

\begin{figure}[!ht]
  \centering
  \includegraphics[width=160mm]{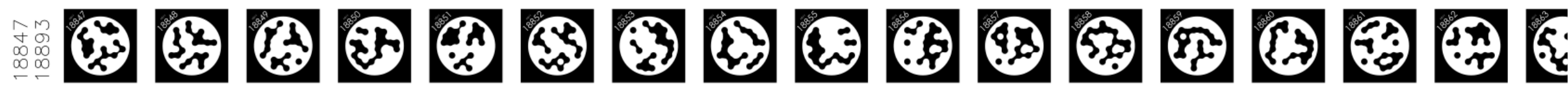}
  \caption{Close-up of a tag stripe. Each fiducial marker features a unique pattern that makes it unequivocally identifiable. Each stripe measures approximately 1 m in length, can hold 47 tags, and contributes to 474 unique stripes.}\label{fig:dedvmethod:tagscloseup}
\end{figure}

To address these challenges, we developed and integrated \emph{T-SLAM}~\cite{SettimiTSlam2024}, which is a timber-centric and primarily tag-based \gls{SLAM} system that employs the STag Library HD11~\cite{Benligiray2019}. This fiducial marker library provides 22{'}309 unique tags, which we reduced to 20~x~20~mm for our study (Fig.~\ref{fig:dedvmethod:tagscloseup}). According to previous studies ~\cite{Benligiray2019, Kalaitzakis2021}, these tags have the lowest false-positive detection rate and highest robustness under steep camera angles or partial occlusions, surpassing other widely used options such as ArUco~\cite{GarridoJurado2014, GarridoJurado2016}, ARTag~\cite{Fiala2005}, ARToolKitPlus~\cite{Wagner2007}, AprilTag~\cite{Wang2016}, and RUNETag~\cite{Bergamasco2011}. T-SLAM handles both the detection and real-scale geometric reconstruction of the timber element and camera localization during fabrication. In T-SLAM, the timber stock is recognized through a two-step process. First, tags are applied to the timber and mapped. The resulting map is then stored and reused during fabrication to enable camera relocalization relative to the timber.

\begin{figure}[!ht]
  \centering
  \includegraphics[width=140mm]{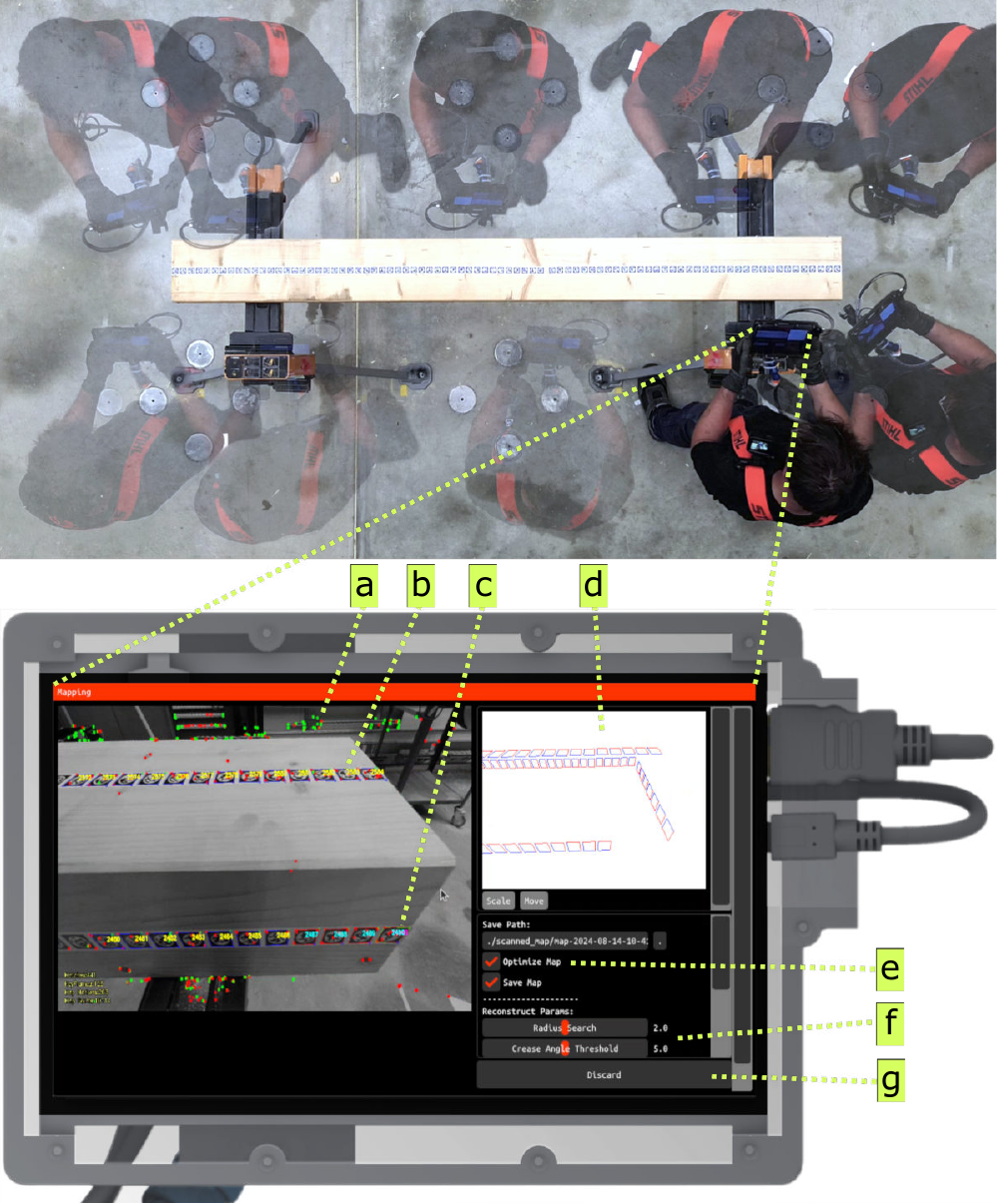}
  \caption{Illustration of the timber mapping procedure in \gls{AC}. The timber piece is flipped and (a) landmark feature points used to facilitate the camera localization, (b) tags already acquired, (c) newly detected tags pending acquisition, (d) map-building overview, (e) optimization parameters, (f) reconstruction parameters for generating a mesh box from the registered tags, and (g) commands to save or discard the current mapping recording.}\label{fig:dedvmethod:mapinit}
\end{figure}

Prior to mapping, one stripe must be applied to each face of the timber beam. During the initial mapping phase (Fig.~\ref{fig:dedvmethod:mapinit}), the environment is scanned to compute the positions of a set of fiducial markers (tags), forming a \emph{tag map} anchored in the \emph{timber frame} \(\{t\}\). Once this tag map is established, it is saved and associated with the timber piece. During the subsequent fabrication phase, the pose of the monocular camera can be determined by matching its current observations of the markers to the known map (Fig.~\ref{fig:dedvmethod:hardover}(b)), yielding the following transformation:

\begin{figure}[!ht]
  \centering
  \includegraphics[width=160mm]{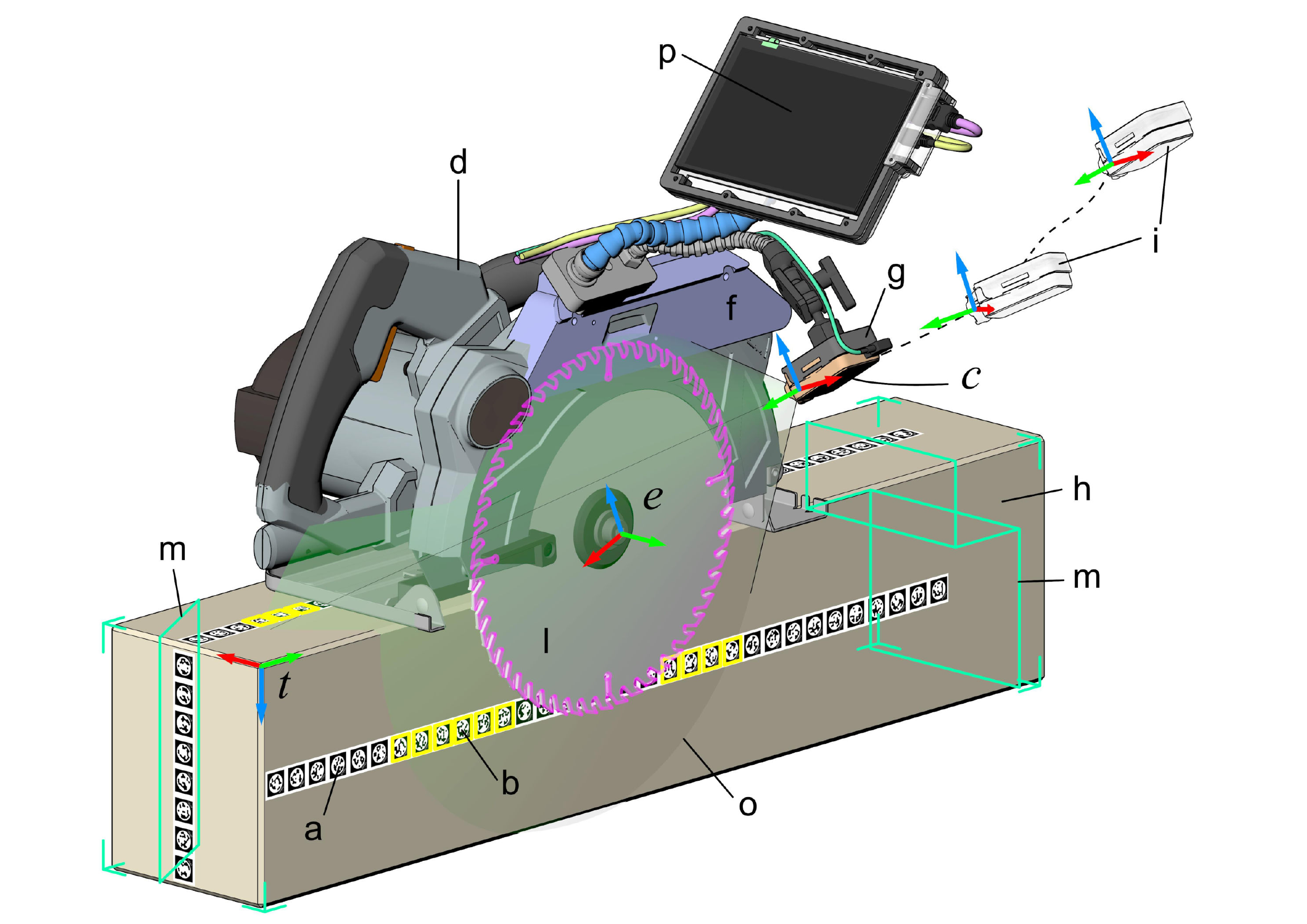}
  \caption{Overview of the \gls{AC} sensing system: (a) tags belonging to the map, (b) detected tags within the camera frame, (d) retrofitted woodworking power tool, (f) 3D mount, (g) monocular camera, (h) timber beam, (i) previous camera-computed poses, (l) toolhead pose detected via its corresponding 3D model contours, (m) execution model registered to the map, (o) fish-eye camera frame boundaries, and (p) touch display .}\label{fig:dedvmethod:hardover}
\end{figure}

\begin{equation}
  \label{equ:gentransform}
  c = {}^{t}T_{\text{t}}^{c}t \; , \; {}^{t}T_{\text{t}}^{c} \in SE(3)
\end{equation}
That is, the pose of the camera with respect to the timber frame.

When the toolhead is detected in the camera images, and its pose is computed in the \emph{camera frame} \(\{c\}\):
\begin{equation}
  e = {}^{c}T_{\text{c}}^{e}c \; , \; {}^{c}T_{\text{c}}^{e} \in SE(3)
\end{equation}
To express the pose of the toolhead in the timber frame, we compose the following transformations:
\begin{equation}
  {}^{t}T_{\text{t}}^{e}
  \;=\;
  {}^{c}T_{\text{c}}^{e}
  \;\cdot\;
  {}^{t}T_{\text{t}}^{c}
\end{equation}
Thus, the \gls{6DoF} pose of the toolhead is consistently referenced to the same tag map that is anchored in the timber frame. By expressing both the camera and toolhead in the timber frame, any beam motion (and its tags) is automatically considered. This \emph{object-based tracking} approach is more robust than methods that rely on background feature points, which can change unpredictably in construction or workshop environments. Consequently, our sensing system remains stable and accurate even when the timber beam is moved or reshaped. In this context, the execution model is added to the 3D scene and registered to the map. The model no longer transforms in space once it is locked to the main timber frame. This is illustrated in Fig.~\ref{fig:dedvmethod:hardover}(b) and (c). if the camera does not detect any map tags in its \gls{FoV}, localization fails and the model is not displayed. However, prior studies on our proposed \gls{SLAM} approach indicate that, with the current tag layout and a $170^\circ$ camera \gls{FoV}, sufficient markers remain in view throughout most cutting and drilling operations~\cite{SettimiTSlam2024}.

\begin{figure}[!ht]
  \centering
  \includegraphics[width=90mm]{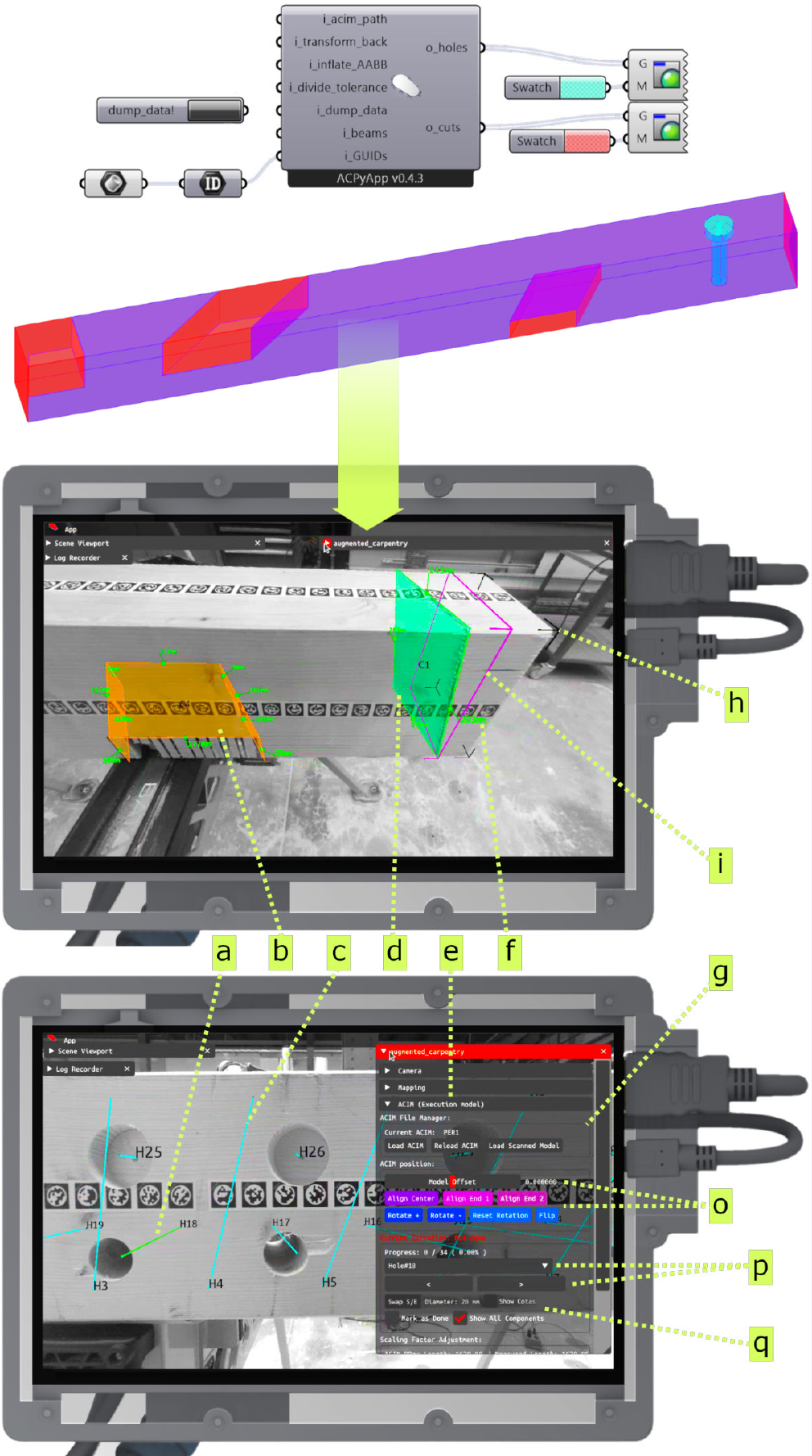}
  \caption{Visualization of the augmented execution model locked to the timber piece: (a) a selected hole, (b) a lap-joint in the unselected state, (c) a hole in the unselected state, (d) a selected lap-joint, (e) a \gls{UI} panel for execution model controls, (f) dimension lines (cotas) that can be toggled on or off, (g) basic I/O controls for \gls{ACIM}, (h) visual cues indicating the bounding box, (i) a widget delineating the outer boundaries of the imported execution model, (o) controls for adjusting the position and orientation of the model, (p) tools to navigate and designate holes or cuts as current, and (q) additional options that are specific to holes or cuts.}\label{fig:dedvmethod:acimview}
\end{figure}

The execution model used by \gls{AC} is imported as an \gls{ACIM} file. Structurally akin to \emph{BTLx}, it is based on an \emph{XML} schema and primarily contains geometric information along with additional fabrication metadata (e.g., fabrication state and ID). During the prototyping phase, we employed Rhinoceros~3D as the primary \gls{CAD} software. The exporter automatically detects all joints and holes via a plugin, requires no manual input, and exports the \gls{ACIM} for each element in a generic 3D model. This allows designers to export a 3D execution format without modifying their primary model for production details or generating supplementary 2D drawings, which are often performed in conventional settings. The \gls{ACIM} metadata can also be updated during fabrication and re-imported by the designer, thereby providing real-time feedback on the execution status.
After the mapping is completed and loaded into \gls{AC}, the user can select the execution model to import. Upon loading, the \gls{AC} parses the geometric information and populates the grayscaled camera feed with colored points, lines, and text labels for the IDs (Fig.~\ref{fig:dedvmethod:acimview}). The system automatically registers the model to the oriented bounding box of the map, aligning it with the timber geometry. The user may refine this alignment along the central axis of the timber to optimize leftover usage. Once the user is satisfied with the augmented positioning, they lock the model in place and can select the next joint to fabricate via the \gls{UI}.

\begin{figure}[!ht]
  \centering
  \includegraphics[width=140mm]{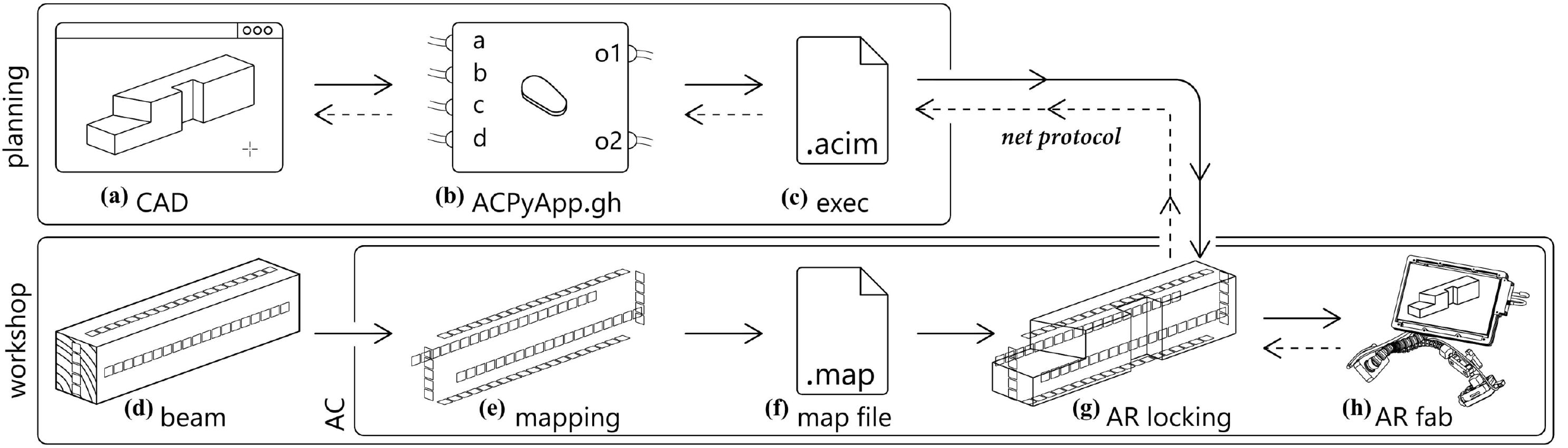}
  \caption{Dataflow for the acquisition and object locking of the beam: (a) the execution model is drawn as a 3D model, and each individual element is exported as (c) an \gls{ACIM} file via (b) a visual scripting plugin. On the workshop side, (d) stickers are applied to the beam, after which (e) the piece is mapped and (f) a new map is generated and stored. Once the \gls{ACIM} file is loaded, (g) object locking is enabled, (h) ensuring that the model is correctly overlaid during fabrication. The dataflow can loop back to the \gls{CAD} station, as the \gls{ACIM} file is continuously updated by the on-site worker to reflect the current fabrication status of the beam.}\label{fig:dedvmethod:dataflow}
\end{figure}

\subsection{Augmented woodworking}
\label{sec::devmethod::woodworking}

\begin{figure}[!ht]
  \centering
  \includegraphics[width=140mm]{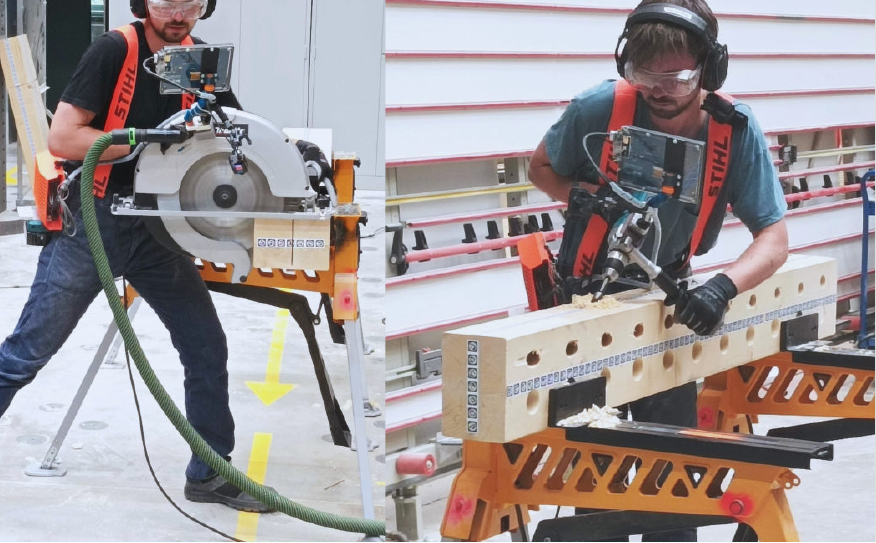}
  \caption{On the left is an augmented woodworking fabrication process incorporating a carpenter saw retrofitted according to the proposed methodology. On the right, a user is shown utilizing augmented feedback during a guided drilling operation.}\label{fig:dedvmethod:faboverview}
\end{figure}

Once the piece is mapped, the execution model is loaded, its overlays are confirmed, and fabrication begins. The proposed framework provides a feedback loop mechanism for cutting and drilling with multiple machines and toolheads to guide the user accurately during the correct execution of the joint via visual cues displayed on the touch monitor (Fig.~\ref{fig:dedvmethod:faboverview}).

The augmented cutting sequence is structured as follows: the user selects the joint to fabricate from the \gls{ACIM} control panel, and if the correct typology of the toolhead is inserted, the visual widgets start to populate the scene as a camera overlay to guide the worker with computed feedback.

\begin{figure}[!ht]
  \centering
  \includegraphics[width=140mm]{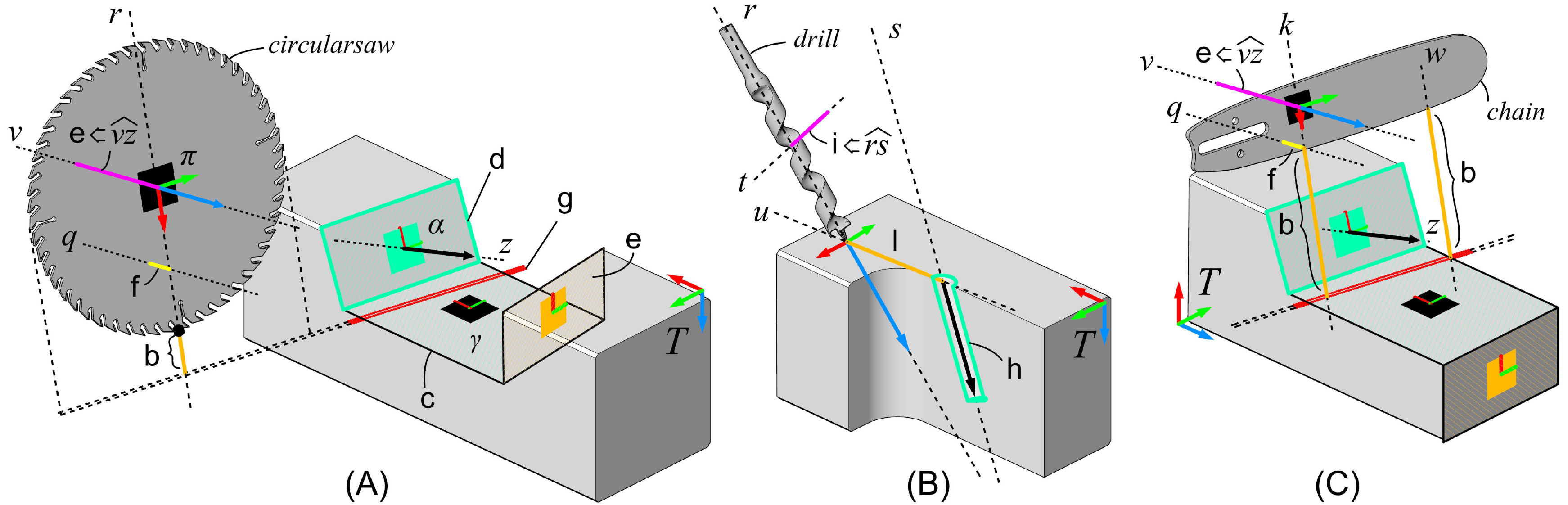}
  \caption{A detailed summary of the geometric computation performed at every frame during fabrication, enabling guidance for the worker using (A) a circular saw, (B) a drill, or (C) a chainsaw.}
  \label{fig:dedvmetghod:fanatomy}
\end{figure}

All user feedback is generated from real-time computations of geometric data. \gls{AC} represents data from the execution model and toolhead as \gls{GO}s, relying on simple geometric primitives instead of complex meshes. While an integrated \gls{6DoF} pose detector uses toolhead library meshes for pose estimation, the system calculations are driven by the \gls{PoI}s defined by the user in the toolhead dataset (Fig.~\ref{fig:dedvmethod:acit}). The distances, angles, and other parameters are computed via live geometric operations between these \gls{PoI}s and the joint geometry in the virtual scene, as shown in Fig.~\ref{fig:dedvmetghod:fanatomy}.

\begin{figure}[!ht]
  \centering
  \includegraphics[width=120mm]{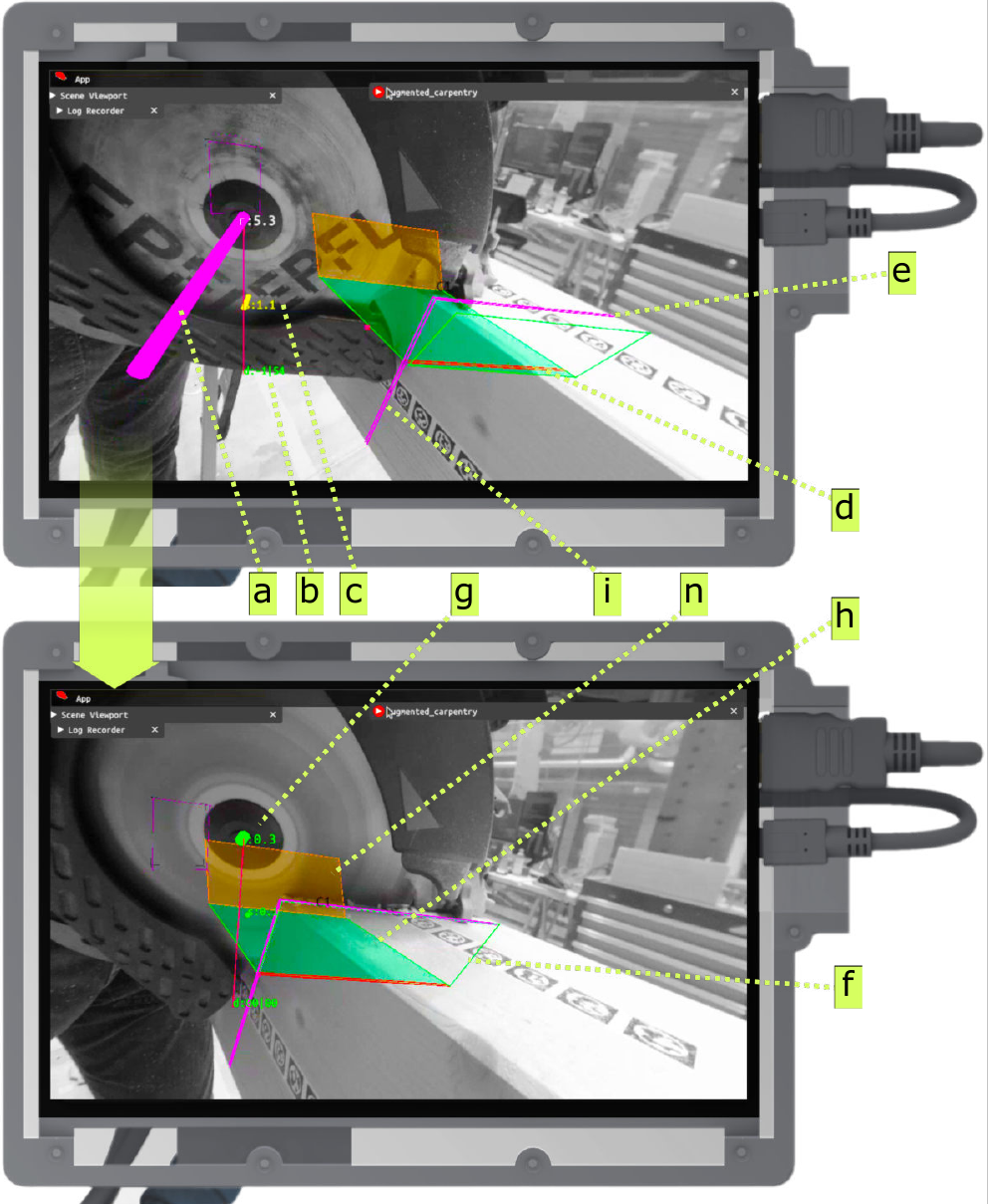}
  \caption{Screenshots of the visual instructions provided by \gls{AC} during cutting. The top row shows the initial positioning, where indicators are not satisfied. The bottom row illustrates the correct alignment, allowing the user to cut. Major \gls{UI} elements include (a) orientation, (b) position, (c) depth guidance, (e-i) blade intersection with the timber section, (d) projection of the blade intersection on the joint, (f) reference face, and (h) adjacent face.}
  \label{fig:dedvmetghod:feedcutclose}
\end{figure}

The current version of the software provides guidance for saw cutting with circular saws, chainsaws, and drill bits, while maintaining a minimal interface to reduce clutter in the \gls{UI}. Each tool requires correct positioning, orientation, and depth, as indicated using a consistent widget design, although they are computed differently for cutting and drilling. During the cutting operation, the system automatically detects the closest face of the joint based on the proximity of the blade, which is highlighted in green (Fig.~\ref{fig:dedvmetghod:feedcutclose}(f)). The user then adjusts the position (Fig.~\ref{fig:dedvmetghod:feedcutclose}(c),~\ref{fig:dedvmetghod:fanatomy}(f)), orientation (Fig.~\ref{fig:dedvmetghod:feedcutclose}(a),~\ref{fig:dedvmetghod:fanatomy}(e)), and depth (that is, blade height; Fig. ~\ref{fig:dedvmetghod:feedcutclose}(b),~\ref{fig:dedvmetghod:fanatomy}(b)) by aligning bars that shift left or right to show the distance from the correct value; each bar also displays a live-updated numerical readout for fine-tuning. The embedded tracking system compensates for the blade thickness (Fig.~\ref{fig:dedvmetghod:feedcutclose}(d),(e)), which facilitates precise positioning—for instance, when cutting a lap-joint by placing the blade within the volume to be removed. The orientation is determined by comparing the normal of the cutting joint face plane with that of the cut plane, allowing for intuitive multi-axis adjustments. Once the position, orientation, and depth values fall within acceptable ranges (highlighted in green in Fig.~\ref{fig:dedvmetghod:feedcutclose}(g)), the user can begin cutting, with the feedback remaining robust despite blade rotation, vibration, timber dust, and other noises. The chainsaw cutting follows the same principle, except that two indicators instead of one are used to track the farthest and bottom points of the elliptical chainsaw bar, ensuring a straight cut even with this generally less precise tool.

\begin{figure}[!ht]
  \centering
  \includegraphics[width=120mm]{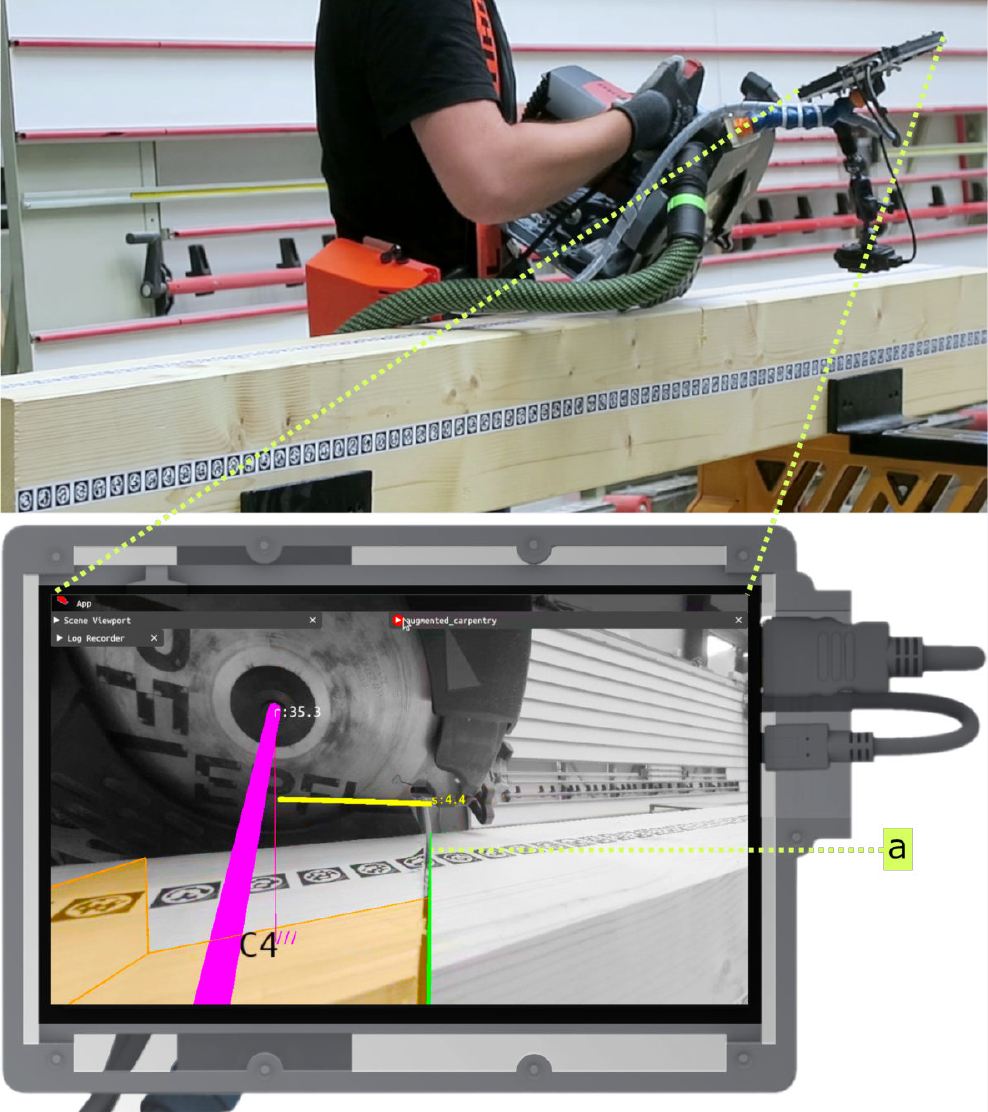}
  \caption{The user visually checks whether the physical cut has been executed correctly by comparing it with the execution model overlay. It can be observed from (a) that the joint side is correctly placed on the right of the cut slot.}
  \label{fig:dedvmethod:prescloseup}
\end{figure}

At each step of the guided cutting process, the worker can verify the precision of the cut by positioning the camera to obtain a clear view of whether the physical slot aligns with the overlaid model (Fig.~\ref{fig:dedvmethod:prescloseup}). Because the sensing detection of \gls{AC} is highly accurate, verifying the correct execution requires only a brief visual inspection, which makes the process straightforward and intuitive.

\begin{figure}[!ht]
  \centering
  \includegraphics[width=120mm]{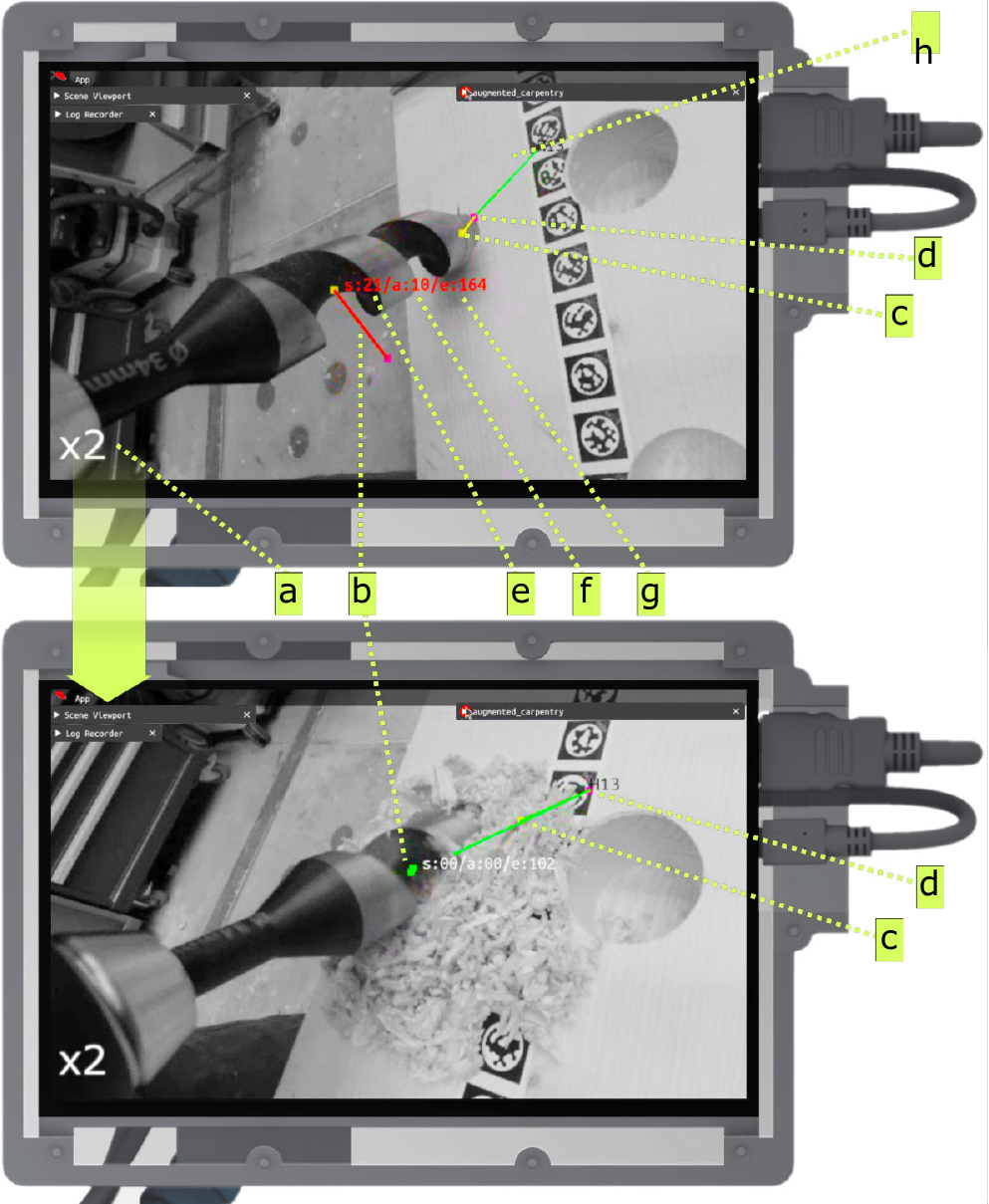}
  \caption{Illustration of the guidance provided by \gls{AC} throughout the drilling procedure. In the top row, the user orientation does not yet satisfy the alignment requirements, while the bottom row demonstrates proper orientation achieved mid-process. The bottom row illustrates the correct alignment during mid-fabrication: (a) the feed is zoomed twice, (b) orientation visual guide, where the correct starting positioning is a line connecting the (c) tip of the toolhead, and (d) the start of the drill hole, with numerical values for the (e) rotation, (f) position, and (g) depth.}\label{fig:dedvmethod:feeddrillclose}
\end{figure}

In drilling scenarios, user feedback is simplified by focusing on two primary vectors: the drill bit and hole axis. By comparing these vectors, \gls{AC} guides the user alignment. Specifically, the system displays two bars: one indicating the orientation, which graphically represents the angle error between the drill bit and hole, as shown in Fig.~\ref{fig:dedvmethod:feeddrillclose}(b),~\ref{fig:dedvmetghod:fanatomy}(i), and the other indicating the initial positioning in the exact figures. Similar to the cutting guidance, these bars display directional cues and inform the user of the residual misalignment, indicating the adjustment required to reach the correct position. Once the alignment is achieved and all widgets turn green, the user drills until the depth indicator confirms the required depth. As illustrated in Fig.~\ref{fig:dedvmethod:feeddrillclose}, the chip accumulation in the scene does not interfere with the tracking signal because the redundancy of multiple tags in the workspace ensures robust performance in the presence of noise.

\begin{figure}[!ht]
  \centering
  \includegraphics[width=100mm]{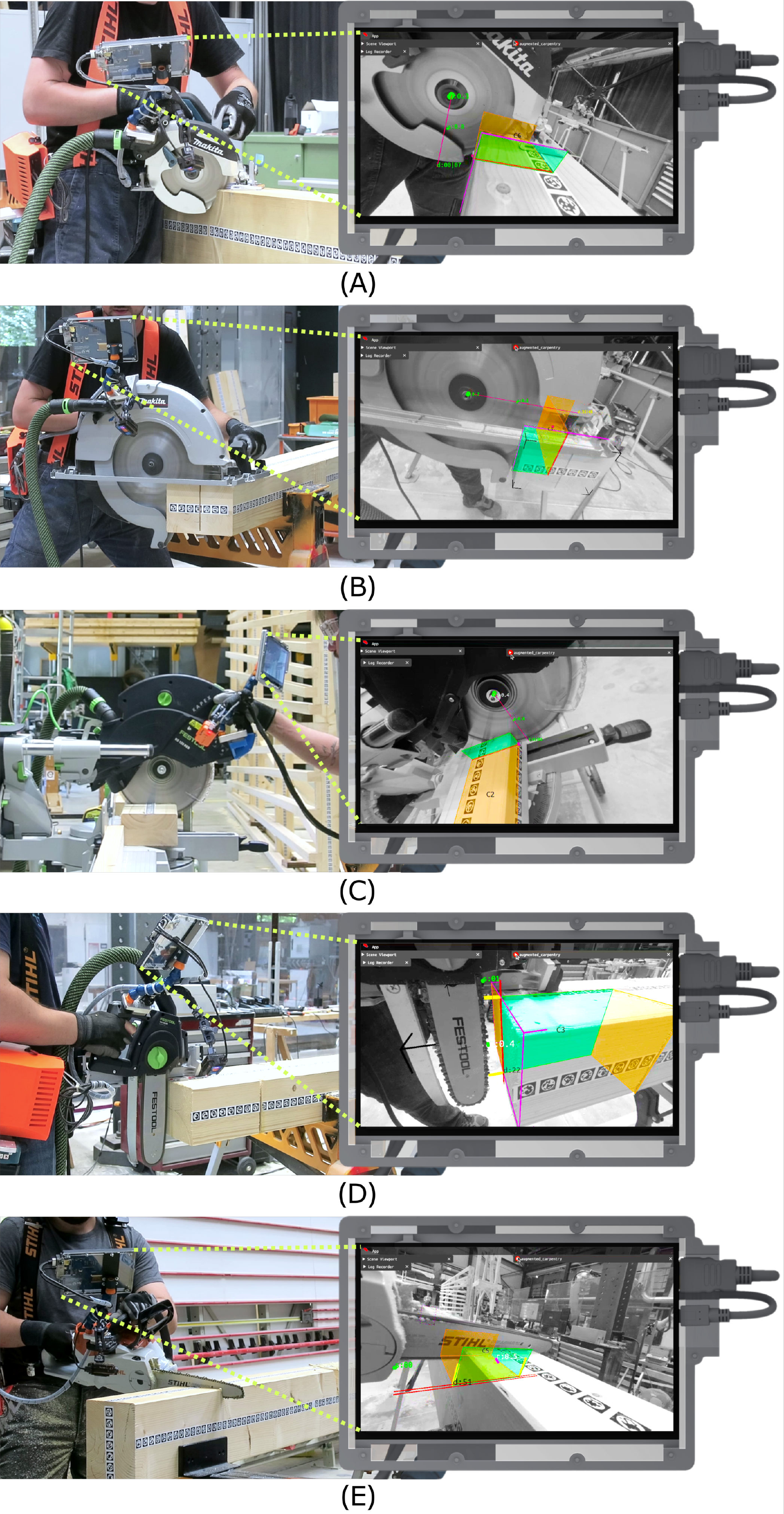}
  \caption{Overview of additional power tools integrated with \gls{AC}: (A) circular saws with (B) 19~mm and 355~mm blade diameters, (C) 260~mm diameter miter saw, (D) sword saw, and (E) chainsaw with a 265~mm length bar.}\label{fig:dedvmethod:ovopec}
\end{figure}

This methodology was validated and implemented using various woodworking power tools. Despite the differences in model and size, each tool employed one of two feedback systems—cutting or drilling—described in this paper. At present, the system supports a miter saw, four types of circular saws with blade sizes ranging from 15 to 35 cm, two electric drills with seven interchangeable bits, and a mid-sized chainsaw, as shown in Fig.~\ref{fig:dedvmethod:ovopec}.

The \gls{AC} feedback system and widget UIs rely on concise geometric visuals and precise numerical data, aligning digital accuracy with the natural dexterity of the craftsperson. We avoid over-automating the process by focusing on live feedback instead of mandating a rigid, step-by-step workflow. In contrast, the system discloses invisible digital information and allows individuals to decide how and when to act, thereby fostering a genuine collaboration between human skills and computational guidance. The proposed minimalistic approach adds computer-like precision and a sense of digital certification to each operation, while preserving operational freedom and reducing the complexity of marking procedures.

\begin{figure}[!ht]
  \centering
  \includegraphics[width=100mm]{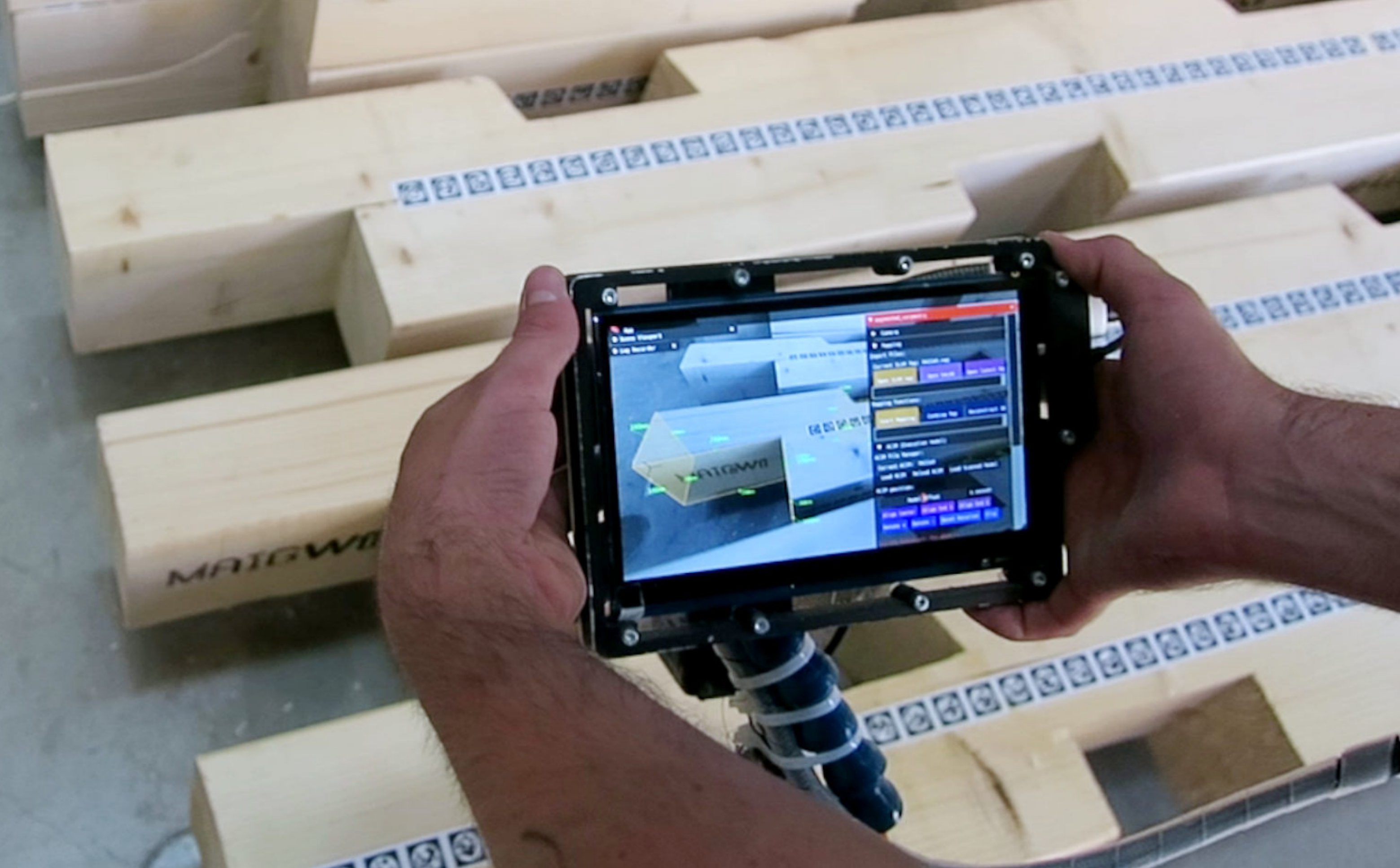}
  \caption{A user inspects a timber piece in a stack, while a graphical overlay displays the corresponding 3D model and dimensions.}\label{fig:dedvmethod:inspect}
\end{figure}

In addition, although it remains in the preliminary stage (see Fig.~\ref{fig:dedvmethod:inspect}), \gls{AC} integrates a real-time monitoring and identification system. The system immediately recognizes each workpiece, thereby verifying and tracking its specifications on the fly. Furthermore, additional information such as measurements, tolerances, or metadata can be displayed live, offering continuous feedback during or after fabrication.

\begin{figure}[!ht]
  \centering
  \includegraphics[width=160mm]{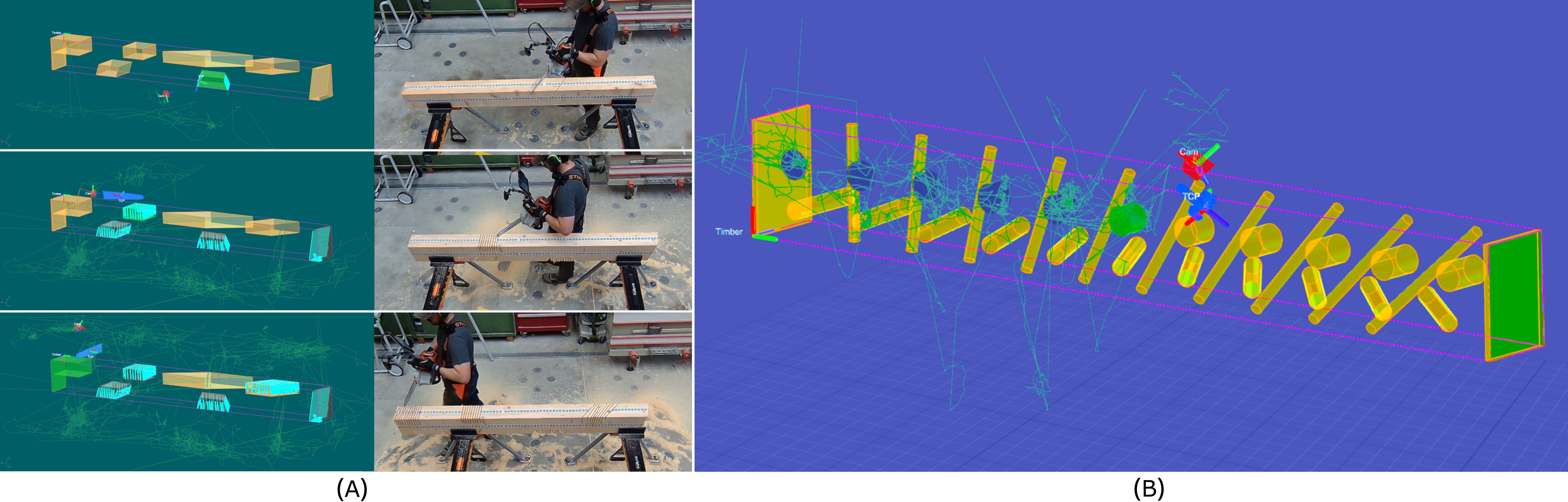}
  \caption3D recorder component of {\gls{AC}: (A) temporal sequences and tracking of an augmented chainsaw operation, in which all cuts are tracked and compiled in the model, and (B) the 3D model generated from the recorded log of drilling operations with multiple drill bits.}\label{fig:dedvmetghod:tdrec}
\end{figure}

Finally, another feature of the \gls{AC} platform is its 3D recording module (Fig.~\ref{fig:dedvmetghod:tdrec}). This component continuously tracks the position and orientation of all relevant physical elements throughout the fabrication process. The system compiles a detailed record of how every component is executed by logging each user decision; for example, adjustments to cutting angles, drilling depths, or any other interactive inputs. This dataset can be replayed in a virtual environment, enabling a step-by-step analysis of the entire operation. By offering transparency regarding where and when deviations occur, the system supports timely adjustments and continuous improvement, ultimately enhancing the reliability, traceability, and efficiency of individual tasks and construction workflows.

\section{Evaluation and performance}
\label{sec:eval}

\subsection{Materials and methods}
\label{sec:eval:matandmet}

In the \gls{AC} accuracy evaluation, we aimed to gauge the overall accuracy of the proposed augmented fabrication guidance system. In previous studies, we provided accurate estimates of each component of the proposed methodology. This includes the accuracy of the integrated \gls{SLAM} ~\cite{SettimiTSlam2024} and toolhead 3D pose detector ~\cite{TTool2024}. However, these components have not been integrated into a complete operating system for evaluation in full-scale fabrication scenarios. Therefore, an extensive experimental campaign involving the production of 1:1 timber structures was designed. We measured the error between each as-fabricated specimen and its corresponding 3D execution model, and treated this measurement as a comprehensive evaluation of the proposed \gls{CV}-guided system. A three-month experimental campaign was conducted by a two-person team consisting of an inexperienced trainee and a mid-level carpenter.

\subsubsection{Specimens}
\label{sec:eval:spec}

\begin{figure}[!ht]
  \centering
  \includegraphics[width=100mm]{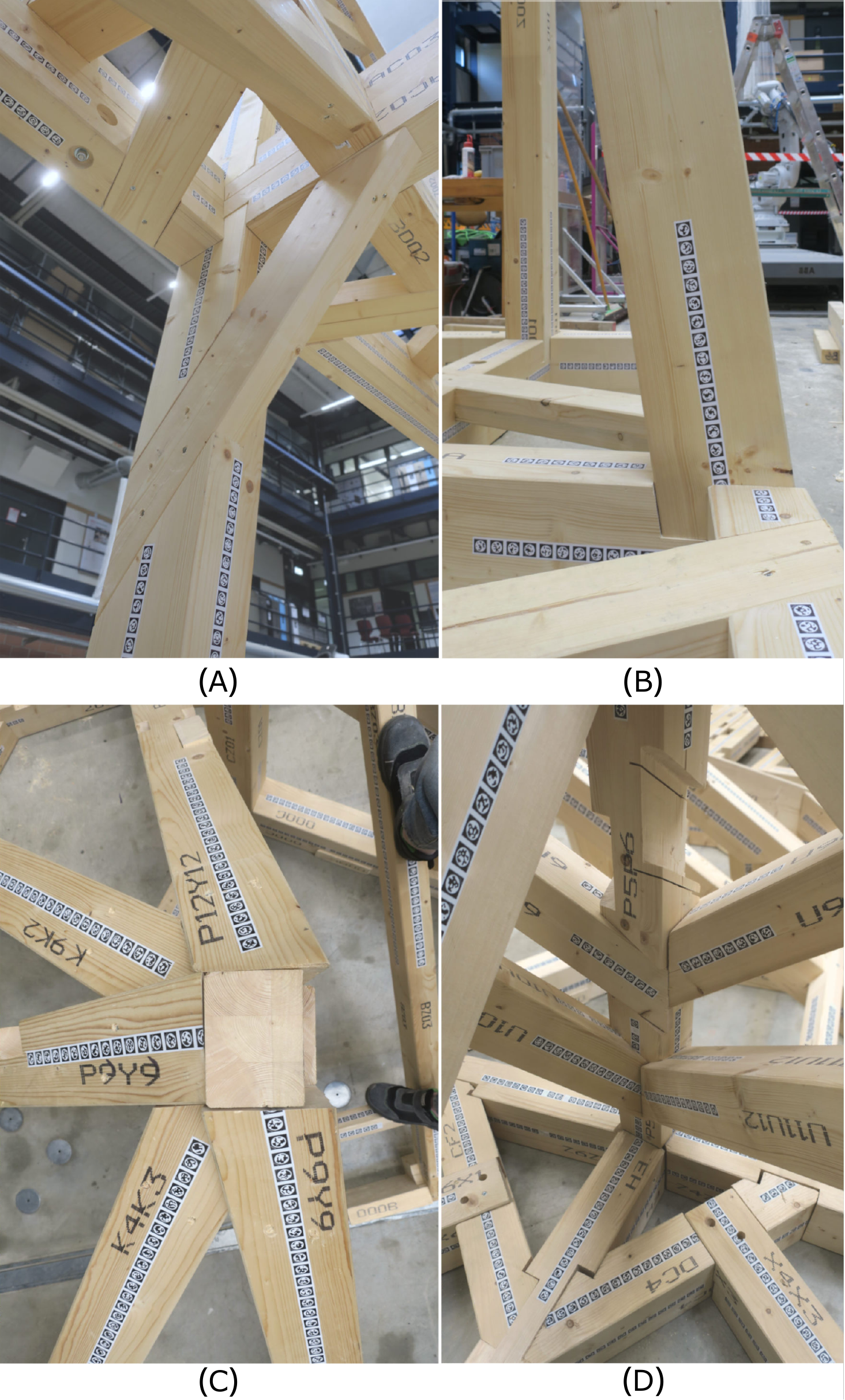}
  \caption{Selection of connections and fabricated joint details: (A) details of the mid-connections and spandrels, (B) floor base slab, (C) top view of the half-roof gable, and (D) the crown post of the triangular roof frame.}\label{fig:eval:detailspnbroof}
\end{figure}

\begin{figure}[!ht]
  \centering
  \includegraphics[width=120mm]{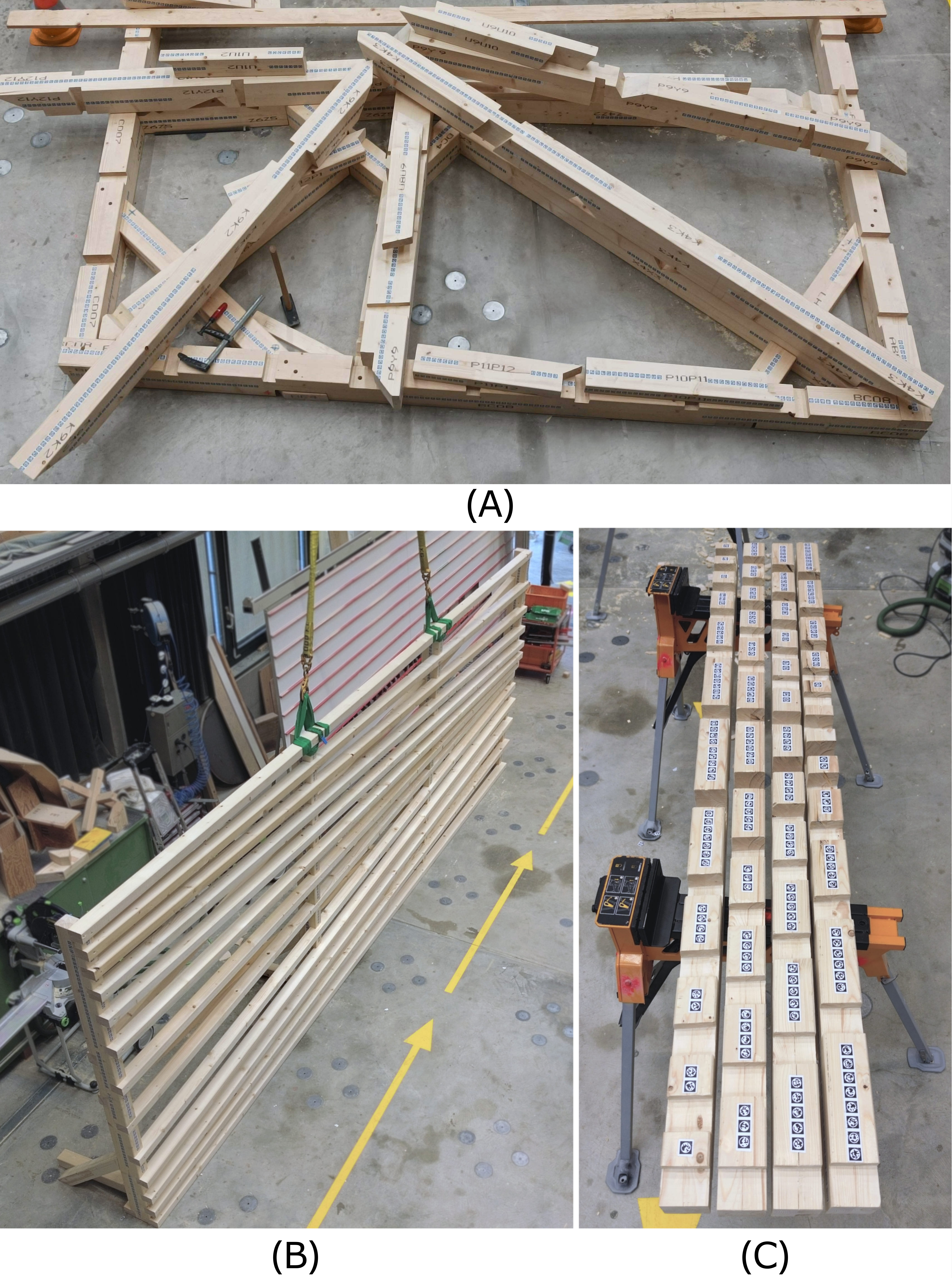}
  \caption{Selection of fabricated mock-ups: (A) top view of a roof structure with an irregular design, the elements of which were fabricated with \gls{AC} and stored in place, ready to be assembled, (B) parametrically designed facade of 6~m length featuring a double-curvature pattern, and (C) a close-up view of its four studs crafted with \gls{AC}.}\label{fig:eval:asblroof}
\end{figure}

In total, 166 joints were executed using the proposed \gls{AC} guidance system on 57 different white spruce elements with lengths ranging from 0.5 to 3.8~m and calibrated cross-sections from 0.14 up to 0.24~m. These realistic lengths and cross-sections are crucial for testing \gls{CV} systems, as such technologies often encounter numerous challenges, potentially leading to critical failures when managing large sets of elements, while ensuring a robust and consistent signal during fabrication. All mock-ups and their components were fabricated over a four-month span within a shop workspace, where we reported a mean temperature of 24.4 $\pm$ 4\degree and humidity of 54.3 $\pm$ 11~$\%$. The produced elements were manually assembled into five large-scale timber structures, incorporating various building systems, from traditional roof structures (Fig.~\ref{fig:eval:assemblydiff}(B.1, C.1)), and post-and-beam frames (Fig.~\ref{fig:eval:assemblydiff}(A.1)) to clad façades (Fig.~\ref{fig:eval:assemblydiff}(D.1)), to test the augmented fabrication system in realistic timber construction scenarios. To ensure comprehensive coverage and provide a rigorous evaluation of the proposed system, we designed a wide range of timber joints that represent both longstanding traditions and widely adopted modern solutions, such as scarf, butt, lap-, or half-lap-joints (Fig.~\ref{fig:eval:detailspnbroof}). Furthermore, all timber elements were parametrized and digitally designed with non-standard cutting angles, with an average angle variance of 90 $\pm$ 180\degree and unique localization within the beam, creating complex designs. Every element was designed with unique angle and distance references, emphasizing the ability of \gls{AC} to assist workers in fabrication tasks that would otherwise require extensive measurement and marking (Fig.~\ref{fig:eval:asblroof}).

\begin{figure}[!ht]
  \centering
  \includegraphics[width=100mm]{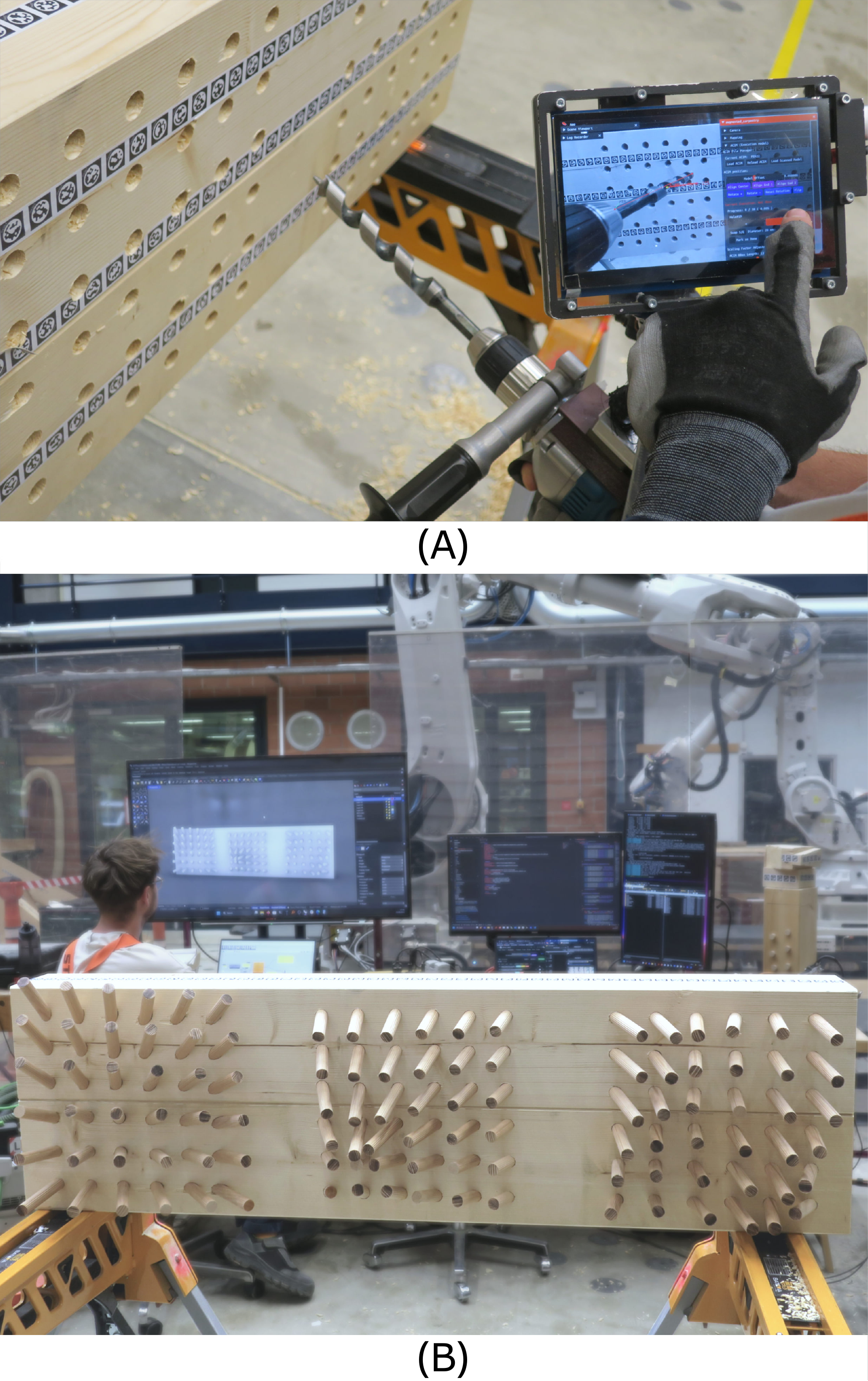}
  \caption{A probing plate was designed specifically for the evaluation of \gls{AC}-guided drilling tasks: {(A)} shows the \gls{AC} interface during the fabrication session, and {(B)} to detect the piercing axis of each hole from the scanned point cloud, a numerically cut hardwood dowel is inserted in each hole. The associated portion of the point cloud for each dowel is then used for comparison with the corresponding \gls{CAD} model.}\label{fig:eval:drillinprobe}
\end{figure}

Drilling is a fundamental task in carpentry for joinery, assembly, and structural connections. A total of 108 holes were drilled into the probing plate, with angles ranging from 30 to 60~\degree{} and depths from 80 to 170~mm, to assess the accuracy of the proposed system. The plate was specifically designed to evaluate the drilling accuracy of AC. It features a complex digital pattern that prevents the user from relying on references other than those provided by the AC interface (see Fig.~\ref{fig:eval:drillinprobe}).

\subsubsection{Augmented Fabrication Tools}
\label{sec:eval:augseltools}

For saw-cutting and drilling, we integrated a set of four tools into \gls{AC}, representing a typical carpentry range:

\begin{enumerate}
  \item \textit{Standard saw-cutting:} For average-sized cuts, we use the Mafell\textsuperscript{\tiny\textregistered} KSP85 and K85E saws. Both are fitted with a 237 mm blade with a maximum cutting depth of 75 mm.
  \item \textit{Large end cuts:} For deeper cuts, we employed the Makita\textsuperscript{\tiny\textregistered} 5143R, which features a 355~mm blade and can cut up to 135~mm in depth.
  \item \textit{Drilling tasks:} For precise wood drilling, we used the Makita\textsuperscript{\tiny\textregistered} DP400. It is equipped with a wood-specific drill featuring a single tracer, a threaded tip, a 20~mm diameter, a seamless cylindrical shaft, and an overall length of 200~mm.
\end{enumerate}

\subsubsection{Measurements methodology}
\label{sec:eval:measmetho}

\begin{figure}[!ht]
  \centering
  \includegraphics[width=160mm]{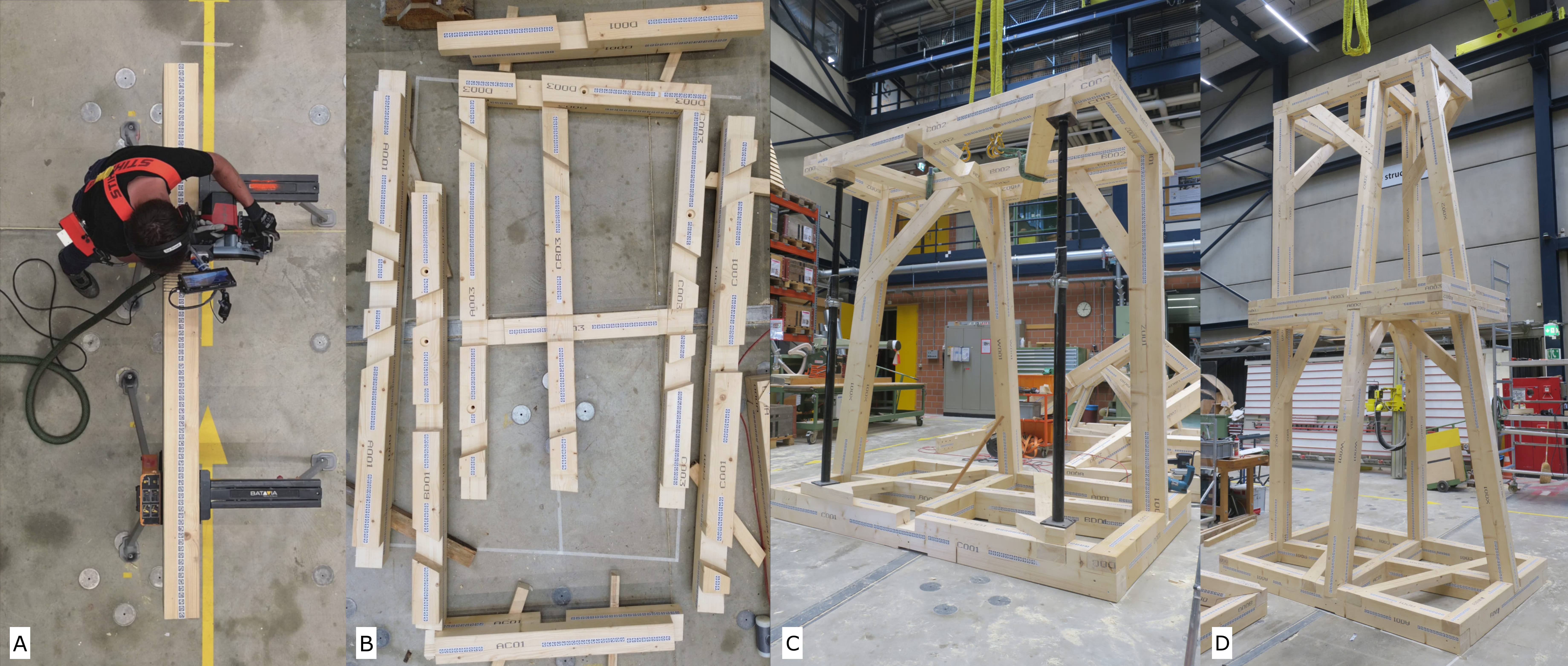}
  \caption{Illustration of the evaluation campaign sequence for the two-story mock-up: (A) each element is initially fabricated with \gls{AC}, (B) scanned individually, (C) assembled into submodules, and (D) ultimately combined into complete structures, which are then fully scanned.}\label{fig:eval:assemblyseq}
\end{figure}

\begin{figure}[!ht]
  \centering
  \includegraphics[width=100mm]{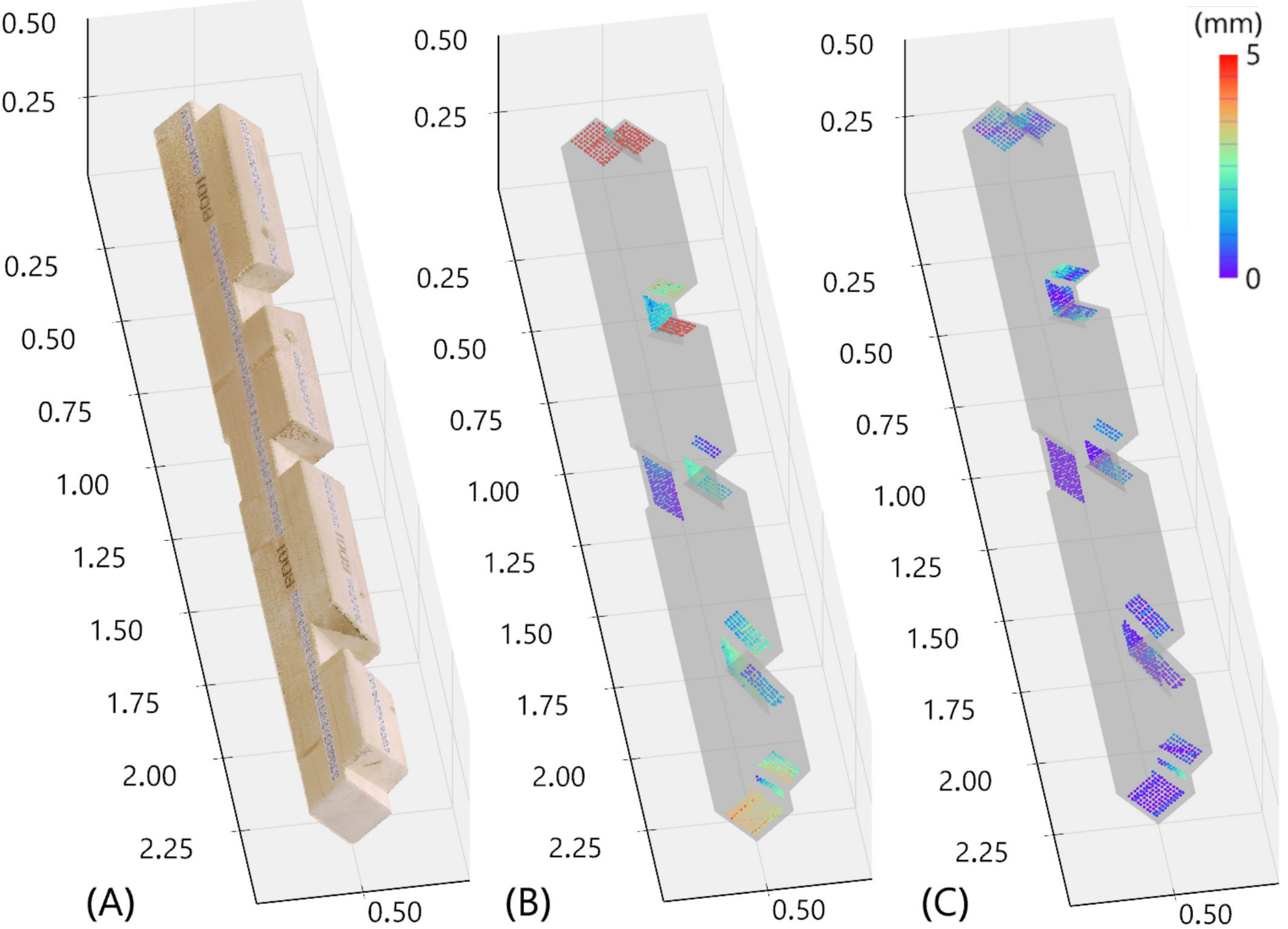}
  \caption{The evaluation methodology of diffCheck for joint analysis on one of the 54 fabricated, scanned, and analyzed elements: (A) shows the captured raw point cloud of the beam. In (B), the scan is registered to its \gls{CAD} model, and the error related to the displacement of each joint is calculated. In (C), the joints are detected in the raw scan and independently registered to their \gls{CAD} counterparts, allowing for an individual fabrication quality assessment of each joint.}\label{fig:eval:beameval}
\end{figure}

\begin{figure}[!ht]
  \centering
  \includegraphics[width=100mm]{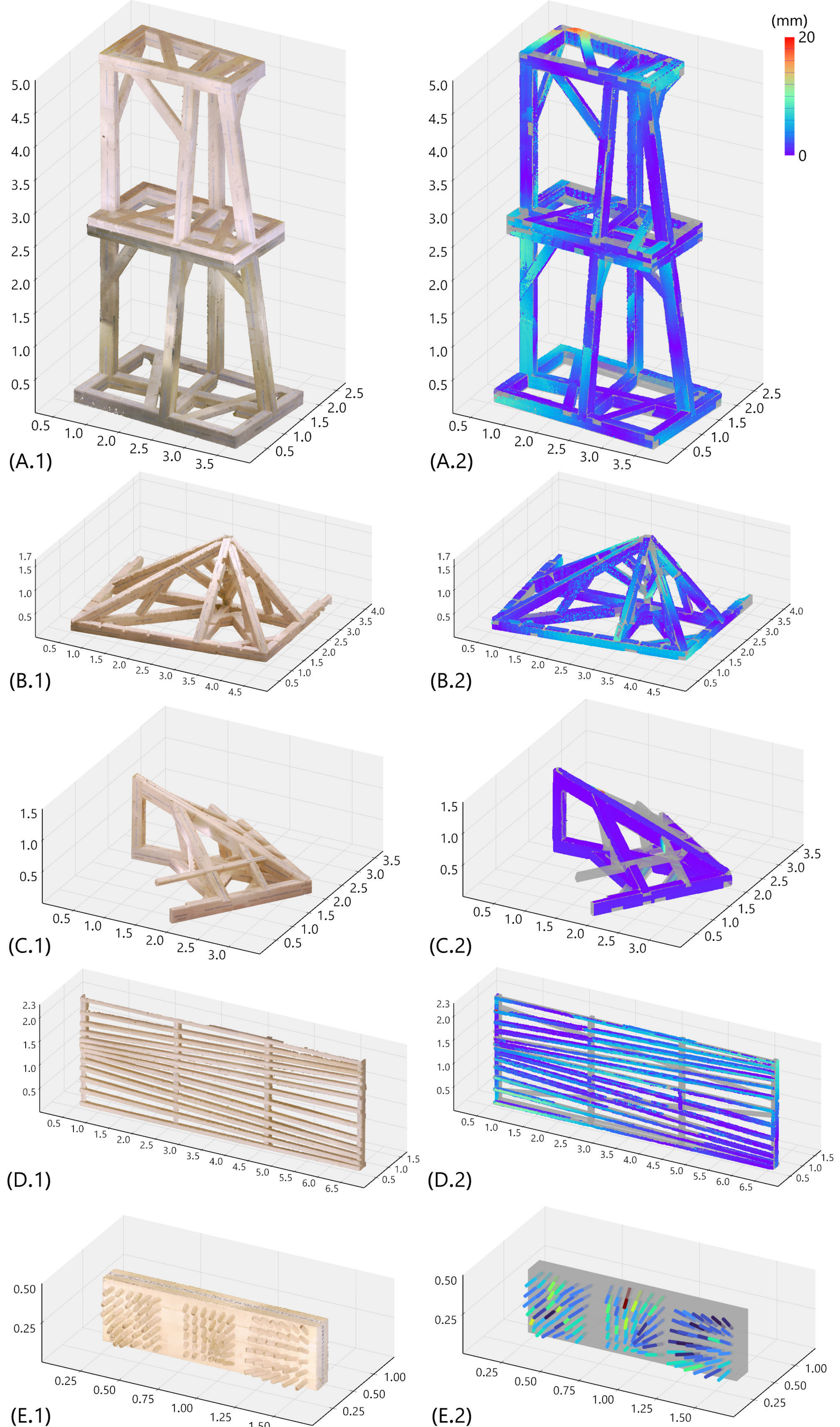}
  \caption{Visualization of deformation on a selected range of assembled timber structures, showing error magnitudes in millimeters. On the left are the raw point clouds of the scans and on the right are the point clouds colored following the distribution error. These include the (A) multi-story timber frame structure, (B, C) triangular truss systems, (D) parametric facade cladding, and (E) drilling probe plate. We omitted the corrupted data in the distance representation and signified it with its corresponding gray \gls{CAD} component.}\label{fig:eval:assemblydiff}
\end{figure}

To measure the discrepancy between the as-fabricated element and its \gls{CAD} model, we acquired the scans of both individual beams (Fig.~\ref{fig:eval:assemblyseq}(B),~\ref{fig:eval:beameval}(A)) and the assembled structures (Fig.~\ref{fig:eval:assemblyseq}(D),~\ref{fig:eval:assemblydiff}(*.1)) using the FARO\textsuperscript{\tiny\textregistered} FreeStyle2 hand scanner, which provides submillimeter accuracy and high-resolution data when capturing at distances under 1.5~m.

We employed \emph{diffCheck}~\cite{diffCheckSoftware2024}, which is free and open-source C\texttt{++}/Python software integrated into Grasshopper, to compare the scans of the fabricated timber structures with their respective \gls{CAD} models. In particular, \emph{diffCheck} registers the scan and model of each timber element, detects and segments the joints, and provides two main outputs. First, the distance error for each joint is calculated, which indicates the positional accuracy of the joint within the piece, as shown in Fig.~\ref{fig:eval:beameval}(B), reflecting the ability of the system to self-locate the piece and its fabrication components. Second, it measures the distance error for each joint face relative to its 3D counterpart (Fig.~\ref{fig:eval:beameval}(C)), revealing how effectively the system guides the precise execution of each joint, and thus provides insight into the overall joint quality. It is important to note that although one joint may be misaligned within the beam, the fabrication instructions for its faces can still be displayed accurately by \gls{AC}, resulting in very low tolerances in the joint quality assessment of the proposed metrics.

Our main objective was to assess the accuracy of each piece to gauge the capacity of the \gls{AC} for localizing, guiding, and correcting the fabrication of joints and holes. In addition, by registering the raw scan data against the 3D model of the final assembled mockup (Fig.~\ref{fig:eval:assemblydiff}(.2)), we investigated whether the cumulative tolerances of individual pieces could eventually render the assembly process infeasible.

\subsection{Results and discussions}
\label{sec:eval:results}

\begin{figure}[!ht]
  \centering
  \includegraphics[width=160mm]{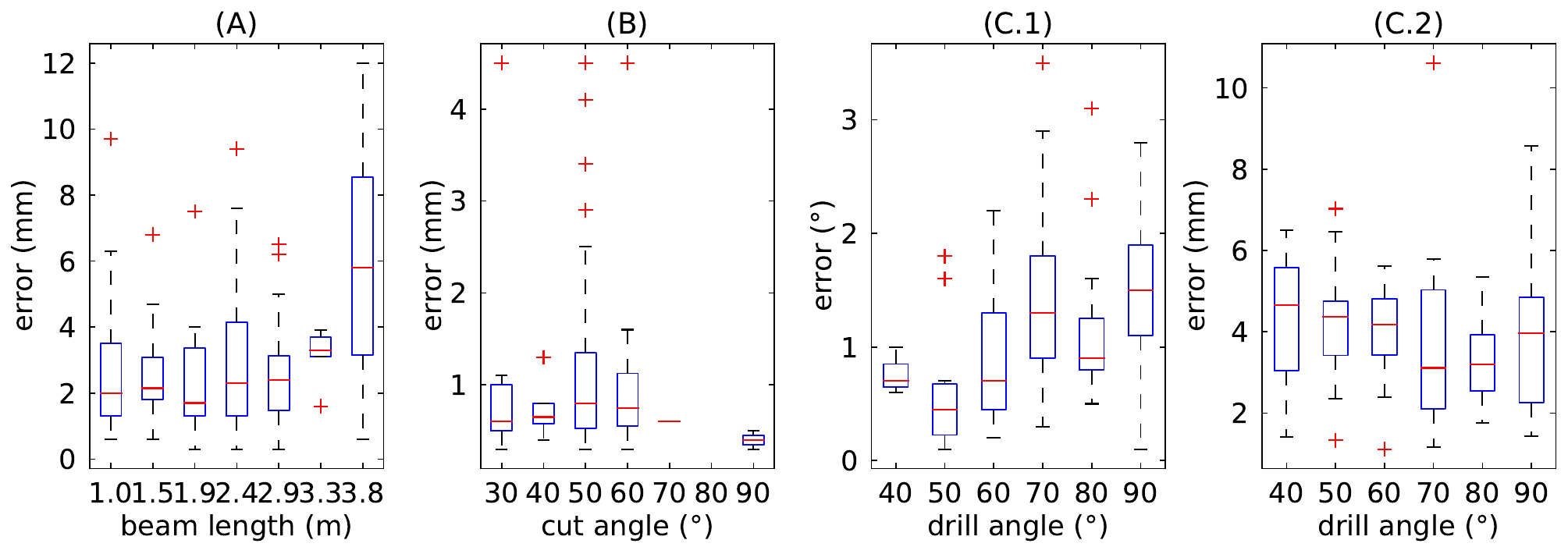}
  \caption{Boxplots of the error for each metric: (a ) error of the position of the joint, (b) quality of the joint execution itself, (c.1) drilling orientation, and (c.2) starting point errors. The central mark on each box represents the median, while the lower and upper edges correspond to the 25th and 75th percentiles, respectively, defining the interquartile range. Whiskers extend to the furthest data points within the non-outlier range and any outliers are distinctly indicated with the "+" symbol.}\label{fig:eval:boxplots}
\end{figure}

This section provides a detailed analysis of the errors associated with each parameter and summarizes their mean values and standard deviations, as presented in Table~\ref{tab:eval:resume}. To illustrate the variability in the joint analysis further, Fig.~\ref{fig:eval:boxplots} show boxplots depicting the distribution of the absolute errors for each parameter. In addition, we provide a series of 3D plots illustrating the assembly deviation for each of the assembled mock-ups in Fig.~\ref{fig:eval:assemblydiff}. Comprehensive experimental data and the corresponding statistical analyses can be accessed through an open-access repository~\cite{AcSupplementary2024}.

\begin{table}[htbp]
  \centering
  \caption{Computed errors for sawing (i.e. joint) and drilling (i.e. perforation) experimental parameters.}
  \renewcommand{\arraystretch}{0.9}
  \begin{tabular}{c|c c|c c|c c c}
    \multicolumn{4}{c}{Joint} & \multicolumn{3}{c}{\multirow{2}{*}{Perforation}} \\
    \cmidrule(r){1-4}
    \multicolumn{2}{c}{Location} & \multicolumn{2}{c}{Faces} & \multicolumn{3}{c}{} \\
    \cmidrule(r){1-2} \cmidrule(r){3-4} \cmidrule(r){5-7}
    Beam   & Distance & Sawing & Distance & Drilling & Orientation & Start position \\
    length & error    & angle  & error    & angle    & error       & error \\
    (\degree)[N\textsuperscript{*}] & (mm) & (\degree)[N] & (mm) & (\degree)[N] & (\degree) & (mm) \\
    \cmidrule(r){1-2} \cmidrule(r){3-4} \cmidrule(r){5-7}
    1.0 [21] & 2.6 $\pm$ 2.1\textsuperscript{**} & 30 [13] & 1.2 $\pm$ 1.5 & -- & -- & -- \\
    1.5 [18] & 2.7 $\pm$ 1.5 & 40 [8] & 0.7 $\pm$ 0.3 & 40 [3] & 0.8 $\pm$ 0.2 & 4.2 $\pm$ 2.6 \\
    1.9 [14] & 2.4 $\pm$ 1.9 & 50 [26] & 1.3 $\pm$ 1.2 & 50 [14] & 0.6 $\pm$ 0.5 & 4.3 $\pm$ 1.6 \\
    2.4 [67] & 2.9 $\pm$ 2.1 & 60 [12] & 1.1 $\pm$ 1.1 & 60 [10] & 0.9 $\pm$ 0.7 & 4.0 $\pm$ 1.4 \\
    2.9 [28] & 2.5 $\pm$ 1.5 & 70 [1] & 0.6 $\pm$ nan & 70 [37] & 1.4 $\pm$ 0.7 & 3.6 $\pm$ 1.9 \\
    3.3 [5] & 3.1 $\pm$ 0.9 & 80 [0] & nan $\pm$ nan & 80 [19] & 1.1 $\pm$ 0.6 & 3.3 $\pm$ 1.1 \\
    3.8 [11] & 6.1 $\pm$ 3.7 & 90 [2] & 0.4 $\pm$ 0.1 & 90 [25] & 1.5 $\pm$ 0.7 & 4.0 $\pm$ 1.9 \\    
    \cmidrule(r){1-2} \cmidrule(r){3-4} \cmidrule(r){5-7}
    \multicolumn{1}{c}{[164]} & 2.9 $\pm$ 2.2 & \multicolumn{1}{c}{[62]} & 0.9 $\pm$ 0.9 & \multicolumn{1}{c}{[108]} & 1.2 $\pm$ 0.7 & 3.8 $\pm$ 1.7 \\
    \multicolumn{7}{r}{(\textsuperscript{*}) {The number of specimens; } \newline (\textsuperscript{**}) {the errors are represented as mean $\pm$ standard deviation.}}
  \end{tabular}
  \label{tab:eval:resume}
\end{table}

In terms of the joint localization, the distance errors indicated a mean value under 3 mm, with standard deviations of 2.2~mm, while presenting a consistent increase with the beam length (Fig.~\ref{fig:eval:boxplots}(A)).
Excluding the initial prototype with beams exceeding 3~m in length, we observed a significant improvement in the joint positioning accuracy, with an average error of 2.3 ± 1.3~mm. We deduce that the proposed methodology provided satisfactory localization and visualization of joints within beams up to 3~m in length. However, this accuracy decreased for beams exceeding this threshold, with the average errors reaching up to 4~m. Such discrepancies are likely owing to corrupted or imprecise mapping of the workpiece. Indeed, achieving reliable results for longer beams proved challenging, often failing. In cases in which a complete map was generated for such lengths, it was frequently distorted at the ends or corrupted, compromising the precision of camera self-localization during fabrication, and hence, the visualization and feedback of the joint. Future improvements should focus on developing a localization mapping methodology that ensures a consistently low tolerance beyond 3~m. It is important to note that this issue is not specific to \gls{AC}, but represents a pivotal scientific challenge requiring fundamental computer vision research that is applicable and beneficial to all digital construction applications. Nevertheless, with the current software version, it is already feasible to assist in the production of 3 m × 3 m spaces in workshops, providing sufficient coverage for most types of applications, particularly in SMEs.

The distance errors for the joint faces were the smallest, averaging below 1~mm (Fig.~\ref{fig:eval:boxplots}(B)). This demonstrates the ability of the proposed software to guide users in cuts for each joint face consistently with submillimeter accuracy, regardless of the beam length or cutting angle.

In contrast, the perforation errors demonstrated a slightly broader distribution and tolerance (Fig.~\ref{fig:eval:assemblydiff}(E)), with the drilling angles (Fig.~\ref{fig:eval:boxplots}(C.1)) and starting position errors (Fig.~\ref{fig:eval:boxplots}(C.2)), indicating a higher variability. The higher tolerance in the starting position errors (3.8 ± 1.7~mm) can be attributed to challenges in the object detection phase. In particular, inaccuracies arose during the detection of the 3D pose of the toolhead, particularly for drill bits. Minor variations in the scale or alignment of the scanned model relative to the actual toolhead can result in positioning errors. This effect is more pronounced for drill bits, which exhibit a more intricate geometry within a smaller camera frame region than a circular saw blade, the detection model of which is based on a single flat surface. 
Furthermore, this inaccuracy can be attributed to the locking of the execution model with the physical piece. Specifically, once the 3D model is locked in place via a rigid transformation referenced to the bounding box of the piece, the bending deformations in the mapping are not considered. Hence, the starting points for the holes may lie inside or outside the actual physical beam surface, producing misleading or erroneous user feedback. A potential solution is to redesign the object-locking mechanism to allow for a certain degree of deformability in the 3D model when registering it on the map of the beam.
However, these discrepancies had a minimal impact on the orientation accuracy, with the drilling angle variability remaining low at 0.7~° and a mean accuracy of 1.2~°. These results highlight the need for improvements in 3D model fidelity, either through enhanced measurement methods for the drill bit or by utilizing \gls{CAD} models from manufacturers. Such improvements can increase the reliability of the system for perforation tasks. Nevertheless, it is important to note that the registered average accuracy remained below 4~mm. For many carpentry operations involving piercing, such as bolting, this level of tolerance may be sufficient to account for the current inaccuracies in drilling guidance.

By comparing the as-built geometry of 57 linear elements with their design specifications, we assess the precision of assembly (Fig.~\ref{fig:eval:assemblydiff}). The results indicate an average error of 4.5 ± 3.1~mm, highlighting the system's capability to produce components within acceptable tolerances for manual assembly consistently. This reliability assures that the fabricated pieces can be joined within standard timber construction tolerances.

\section{Conclusions}
\label{sec:conc}

In this study, ordinary woodworking tools were successfully integrated into a multi-object-aware augmented framework to assist in carpentry tasks. We presented an open-source fabrication software, \gls{AC}, and evaluated its technical performance by focusing on challenges, potential bottlenecks, and system accuracy in tracking both objects and tools during the fabrication sequence. The proposed framework enables precise saw-cutting and drilling based on real-time, computer-generated feedback, thereby eliminating the need for traditional 2D execution drawings, markings, and jigs.
An evaluation of the experimental campaign on one-to-one scale mockups demonstrated that the proposed methodology achieved satisfactory results in both joint visualization and fabrication tasks. The average errors were below 3~mm for joint positioning and millimeter accuracy for joint face cuts, ensuring precise assembly for beams up to 3~m in length. However, challenges persist for longer beams owing to mapping inaccuracies, which require further research in CV to enhance the robustness. The perforation tasks exhibited slightly higher variability, particularly in the starting position errors, but remained within tolerable limits for many carpentry operations. Overall, the system consistently delivered components with acceptable tolerances for reliable woodworking and assembly in assisted timber fabrication workflows.
A UI evaluation was not included in this study because we deliberately prioritized the technical accuracy of the sensing system over user interaction at this stage. First, we established a robust foundation for future investigations into user-centric interface aspects by addressing the overall accuracy of the system. This level of precision is crucial because it ensures that further research on UIs is grounded in clearly quantified accuracy metrics. This factor is of great importance to user experience because an accurate data display is essential.

This study illustrates how manual operations can seamlessly transition into the digital value chain. A significant shortcoming of the traditional \gls{BIM} approach is its failure to incorporate on-site craft and labor operations, which constitute a substantial part of any construction project. By integrating human laborers into a dynamic 3D digital environment that unifies the execution, fabrication instructions, and constraints, \textit{AC} enhances human agency. Rather than reducing the role of skilled labor, this augmented system empowers workers to make well-informed decisions from a broader project perspective. The result is a transformative vision for carpentry that allows workers to contribute to the digital communication chain with ease.

In our design, the system reacts to the actions of the user, rather than dictating a fixed workflow. Crucially, instead of sequencing tasks, our system continuously monitors and quantifies deviations using precise metrics linked to specific movements or actions. This transparent error feedback allows operators to understand the nature and magnitude of any discrepancies, thereby enabling real-time adjustments. By converting errors into actionable insights, our dynamic feedback mechanism extends traditional methods to a responsive, human-centric technological framework.

Sustainable construction requires sustainable production methods. By reconsidering the role of embedded digital tools, we demonstrate how to create systems that are both technologically advanced and widely accessible. Ultimately, retrofitting represents a pragmatic evolution that leverages proven infrastructures to meet modern technological demands while augmenting human agency and fostering a sustainable, distributed, and inclusive digital ecosystem for manual operations in timber construction.

\section{Supplementary materials}
The point clouds from the 3D scans, reconstructed 3D data, and statistical data analysis used in Section~\ref{sec:eval:results} are available in the open-access data repository~\cite{AcSupplementary2024}. The source code~\cite{AugmentedCarpentrySoftware2024} and online documentation have also been released under a permissive license and shared with the entire community.

\section{Credit authorship and contribution statement}
Methodology: A.S.; software: A.S. and contributors; validation: A.S. and J.G.; formal analysis: A.S.; investigation: A.S.; resources: Y.W.; data curation: A.S.; writing of original draft preparation: A.S.; writing of review and editing: A.S. and J.G. and Y.W.; visualization: A.S.; supervision: J.G. and Y.W.; project administration: Y.W.; funding acquisition: Y.W. All authors have read and agreed to the published version of the manuscript.

\section{Acknowledgments}
The authors gratefully acknowledge the financial support provided by the École Polytechnique Fédérale de Lausanne (EPFL). We extend our sincere thanks to the Structural Engineering Group for their invaluable assistance in the design and fabrication of the 3D printed mounts, and particularly to the group's roboticist Gregory Spirlet, metal fabrication technician Krkic Armin, and carpenter technician François Perrin for their expert guidance on timber-related matters. We also appreciate the support and assistance of the Center for Imaging at EPFL, notably Dr. Edward Andò and Dr. Florian Aymanns. Furthermore, we acknowledge the software developers who contributed to the \textit{Augmented Carpentry} project, including Hong-Bin Yang, Naravich Chutisilp, Nazgul Zholmagambetova, and Nicolas Richard, from the SCITAS center at EPFL. Special thanks are also due to Arthur Billotte for his direct contributions to the \textit{Augmented Carpentry} experimental campaign.

\printglossary[type=\acronymtype]

\bibliographystyle{elsarticle-harv} 
\bibliography{cas-refs.bib}

\end{document}